\def\year{2021}\relax
\documentclass[letterpaper]{article} 
\usepackage{aaai21}  
\usepackage{times}  
\usepackage{helvet} 
\usepackage{courier}  
\usepackage[hyphens]{url}  
\usepackage{graphicx} 
\usepackage{natbib}  
\usepackage{caption} 
\usepackage{amsmath}
\usepackage{amssymb}
\usepackage{xcolor}

\usepackage{boldline}
\usepackage{multirow}
\usepackage{hhline}
\usepackage{subfigure}

\usepackage{dblfloatfix}
\usepackage{rotating}

\usepackage[switch]{lineno}

\definecolor{my_orange}{RGB}{255, 165, 0}

\definecolor{my_red}{RGB}{255, 33, 0}

 \title{Learning Intuitive Physics with Multimodal Generative Models}
 \author{
    Sahand Rezaei-Shoshtari,\textsuperscript{\rm 1,\rm 2}
    Francois R. Hogan,\textsuperscript{\rm 1}
    Michael Jenkin,\textsuperscript{\rm 1,\rm 3}\\
    David Meger,\textsuperscript{\rm 1,\rm 2}
    Gregory Dudek\textsuperscript{\rm 1,\rm 2}\\
 }
 \affiliations {
    \textsuperscript{\rm 1} Samsung AI Center Montreal,
    \textsuperscript{\rm 2} McGill University,
    \textsuperscript{\rm 3} York University \\
    \texttt{sahand.rezaei-shoshtari@mail.mcgill.ca, \{f.hogan, greg.dudek\}@samsung.com, \\ \{m.jenkin, david.meger\}@partner.samsung.com}
}

\begin{document}

\maketitle

\begin{abstract}
Predicting the future interaction of objects when they come into contact with their environment is key for  autonomous agents to take intelligent and anticipatory actions. This paper presents a perception framework that fuses  visual and tactile feedback to make predictions about the expected motion of objects in dynamic scenes. Visual information captures object properties such as 3D shape and location, while tactile information provides critical cues about interaction forces and resulting object motion when it makes contact with the environment. Utilizing a novel See-Through-your-Skin (STS) sensor that provides high resolution  multimodal sensing of contact surfaces, our system captures both the visual appearance and the tactile properties of objects. We interpret the dual stream signals from the sensor using a Multimodal Variational Autoencoder (MVAE), allowing us to capture both modalities of contacting objects and to develop a mapping from visual to tactile interaction and vice-versa. Additionally, the perceptual system can be used to infer the outcome of future physical interactions, which we validate through simulated and real-world experiments in which the resting state of an object  is predicted from given initial conditions.
\end{abstract}

\section{Introduction}

Recently, several authors have pointed out the synergies between the senses of touch and vision: one enables direct measurement of 3D surface and inertial properties, while the other provides a holistic view of the projected appearance. Methods such as \citet{li2019connecting} have  trained joint perceptual components, allowing better inference of physical properties from images, for example. This paper extends this reasoning into dynamic prediction: how can we predict the future motion of an object  from visual and tactile measurements of its initial state? If a previously unseen object is dropped into a human's hand, we are able to infer the object's category and guess at some of its physical properties, but the most immediate inference  is whether it will come to rest safely in our palm, or if we need to adjust our grasp on the object to maintain contact. Vision allows rapid indexing to capture overall object properties, while the tactile signal at the point of contact fills in a crucial gap, allowing direct physical reasoning about balance,  contact forces and slippage. This paper shows that the combination of these signals is ideal to predict the object motion that will result in dynamic scenarios. 
Namely, we predict the final stable outcome of passive physical dynamics on objects based on sensing their initial state with touch and vision. 




\begin{figure}[t]
    \centering
    \includegraphics[width=0.35\textwidth]{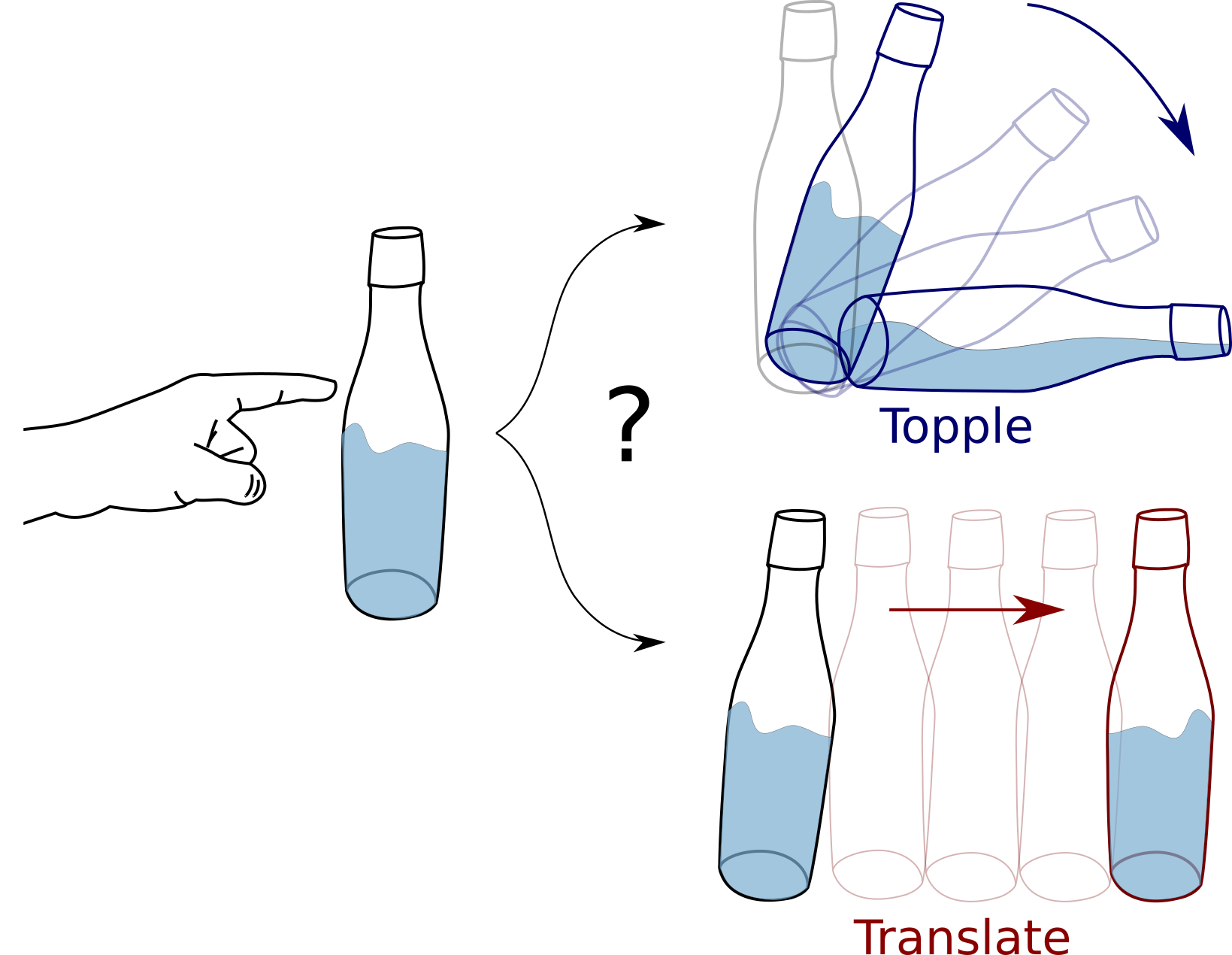}
    \caption{Predicting the outcome of physical interactions. Given an external perturbation on a bottle, how can we predict if the bottle will topple or translate? This paper reasons across visual and tactile modalities to infer the motion of objects in dynamic scenes.}
    \label{fig:bottle_toppling}
\end{figure}

Previous research has shown that it is challenging to predict the  trajectory of objects in motion,  due to the unknown frictional and geometric properties and indeterminate pressure distributions at the interacting surface~\cite{fazeli2020fundamental}. To alleviate these difficulties,  we focus on learning a predictor trained to capture  the most informative and stable elements of a motion trajectory. This is in part inspired by recent findings in Time-Agnostic Prediction \cite{jayaraman2018time}, where the authors show that the prediction accuracy and reliability of predictive models can be vastly improved by focusing on outcomes at key events in the future. Furthermore, this approach mitigates the error drift from which visual prediction typically suffer; where uncertainties and errors propagate forward in time to produce blurry and imprecise predictions.  For example,  in Fig.~\ref{fig:bottle_toppling}, when  predicting the outcome of  an applied push on a bottle, an agent should reason about the most important consequence of this action: will the bottle topple over or will it translate forward?  
To study this problem, we present a novel artificial perception system, composed of both hardware and software contributions, that allows measurement and prediction of the final resting configuration of objects falling on a surface. We  prototype a novel sensor able to simultaneously capture visual images and provides tactile measurements. The data from this new \emph{See-Through-your-Skin} (STS) sensor is interpreted by a multimodal perception system inspired by the multimodal variational autoencoder (MVAE) \cite{wu2018multimodal}.
\begin{figure}[t]
    \centering
    \includegraphics[width=0.4\textwidth]{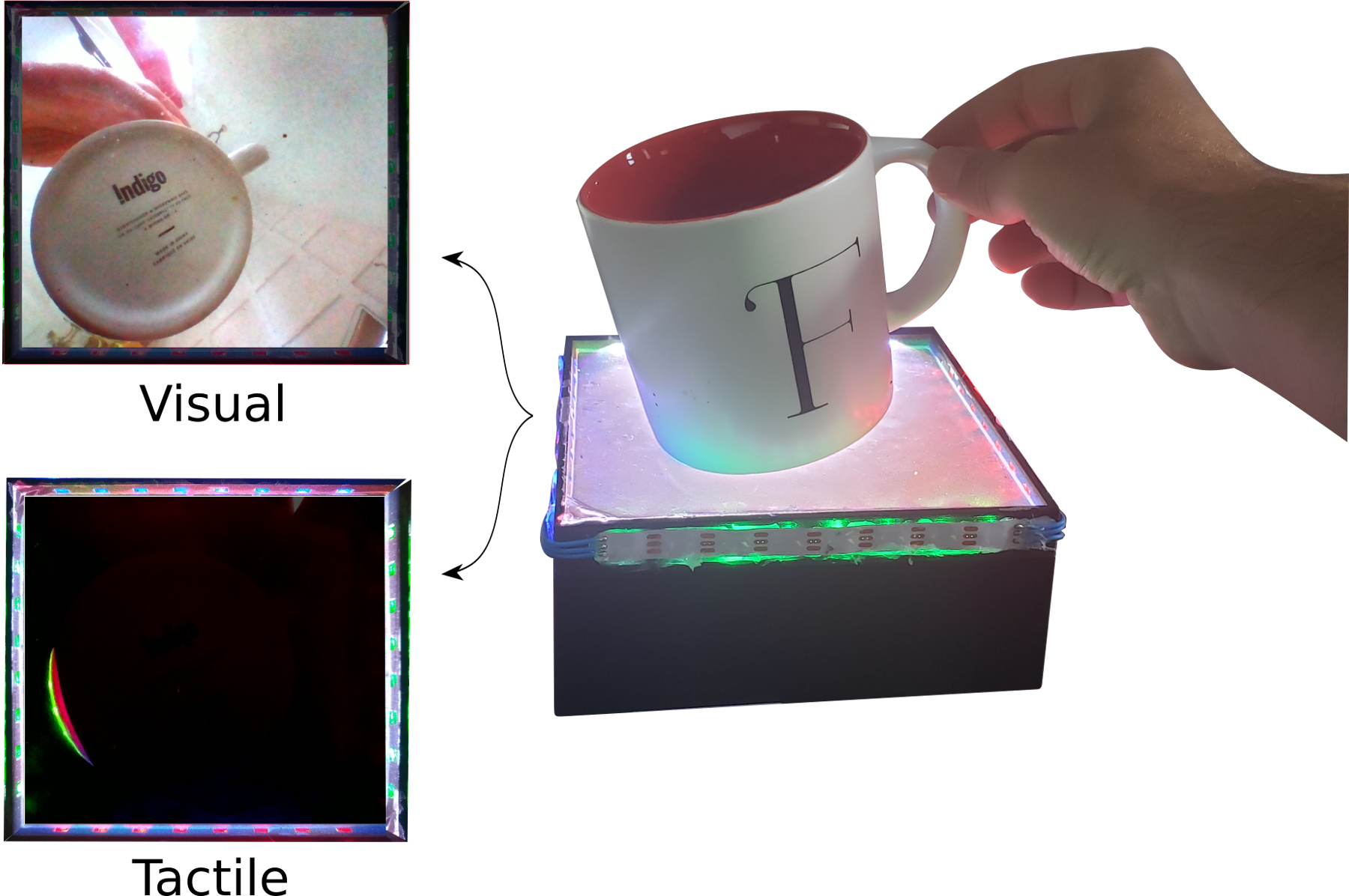}
    \caption{The visuotactile multimodal output of the  See-Through-your-Skin (STS) sensor. Using controlled internal illumination, the surface of the sensor can be made to be transparent, as shown in the top left, allowing the camera to view the outside world.  In the lower left figure, the sensor provides a tactile signature by maintaining the interior of the sensor bright relative to the outside. }
    \label{fig:visuotactile_sensor}
\end{figure}
The main contributions of this paper are:

\textbf{Generative Multimodal Perception}  that integrates visual and tactile feedback within a unified framework. The system, built on Multimodal Variational Autoencoders (MVAE),  exploits visual and/or tactile feedback when available, and learns a shared latent representation that encodes information about object pose, shape, and force to make predictions about object dynamics. To the best of our knowledge, this is the first study that learns multimodal dynamics models using collocated visual and tactile perception.

\textbf{Resting State Predictions} that predicts the resting state of an object during physical interactions. We show that this approach is able to learn generalizable dynamics model capable of learning physical scenarios without explicit consideration of a full physical model. 
 
 \textbf{Visuotactile Dataset} of object motions in dynamic scenes. We consider three scenarios: objects freefalling on a flat surface, objects sliding down an inclined plane, and objects perturbed from their resting pose. Our code is made publicly available \footnote{\scriptsize \texttt{https://github.com/SAIC-MONTREAL/multimodal-dynamics}}.

\section{Related Work}
\subsubsection{Optical Tactile Sensors}  utilize visual information to encode touch-based interactions~\cite{shimonomura2019tactile, Abad:20}. 
Using a combination of a camera and a light source to capture distortions in the contact surface introduced by tactile interactions, optical sensors such as  GelForce~\cite{vlack2005gelforce}, Gelsight~\cite{Johnson:09}, Omnitact~\cite{padmanabha2020omnitact}, DIGIT ~\cite{lambeta2020digit}, GelSlim~\cite{donlon2018gelslim},  and Soft-Bubble~\cite{kuppuswamy2020soft}  render a high resolution image of the contact geometry.  
 These optical tactile sensors make use of an opaque membrane that  obscures the view of the object at the critical moment prior to  contact. This research builds on recent developments to optical sensors enabling them to simultaneously render both tactile and visual feedback from a same point of reference, such as the Semi-transparent Tactile Sensor~\cite{hogan2020seeing} and FingerVision~\cite{yamaguchi2017implementing}.

\subsubsection{Multimodal Learning of Vision and Touch} have been shown to produce rich perceptual systems that exploit the individual strengths of each modality. Multimodal approaches have been proposed for 3D shape reconstruction \cite{Allen1984Surface,Bierbaum2008,Ottenhaus2016,Yi2016Active,17-driess-IROS,Luo2016, wang20183d,luo2018vitac, smith20203d}, pose estimation~\cite{izatt2017tracking}, robotic manipulation  \cite{li2014localization, calandra2017feeling, calandra2018more, Lee2019,watkins2019multi}, and dynamics modeling \cite{tremblay2020multimodal}. The connections between both modalities have been investigated in \cite{li2019connecting, lee2019touching} using GANs and concatenation of embedding vectors from different sensing modes \cite{yuan2017connecting, lee2019making}. The perceptual system proposed in this research  maps vision and touch to a shared latent space in an end-to-end framework, resulting in a robust predictive model capable of handling missing modalities. We focus on learning dynamic models involving objects interacting with their environment rather than explicitly studying the connection between the two modes.

\subsubsection{Temporal Abstraction} in planning and prediction can mitigate the challenges in long-horizon tasks \cite{sutton1999between} by temporally breaking down the trajectories into shorter segments. This can be achieved through the discovery of prediction bottlenecks, shown to be effective at generating intermediate sub-goals \cite{mcgovern2001automatic, bacon2017option}   used by agents in a  hierarchical reinforcement learning paradigm~\cite{nair2019hierarchical, nair2020goal, pertsch2020keyframing}. Recent work on Time-Agnostic Prediction has shown that the notion of predictability can be  exploited to identify  bottlenecks~\cite{jayaraman2018time, neitz2018adaptive}, by skipping frames with high prediction uncertainty and focusing on frames that are more stable and easier to predict. This paper draws inspiration from this work by defining  the state bottlenecks as stable object configurations during dynamic interactions and show that this assumption allows to learn more accurate and robust dynamics models.

\section{Approach}

This section outlines our approach for learning intuitive physical models that reason across visual and tactile sensing. Our long term objectives are twofold. First, we aim to understand how to develop reliable sensory perceptual models that  integrate the senses of touch and sight  to  make inferences about the physical world. Second, we seek to exploit these models to enable autonomous agents to interact with the physical world. This paper focuses on the former, by investigating the core capability of a visuotactile sensor to make predictions about the evolution of dynamic scenarios.

While dynamic prediction is most often formulated as a high resolution temporal problem, we focus on predicting the final outcome of an object's motion in a dynamic scene, rather than predicting the fine-grained object trajectory through space. With an understanding that the main purpose of predictive models is to allow autonomous agents to take appropriate actions by reasoning through the consequences of those actions on the world, we believe that in many scenarios, reasoning about targeted  future events is sufficient to make informed decisions. Importantly, we examine the relevant tactile information to make such predictions.  Whereas motion prediction is most commonly approached as a purely visual product, we highlight the importance of reasoning through physical phenomena such as interaction forces, slippage, contact geometry, etc., to make informed decisions about the object state. 


\section{Visuotactile Sensing}
\label{sec:visuotactile_sensing}

This  section  describes a novel  visuotactile sensor, named the See-Through-your-Skin (STS) sensor,   that renders dual stream high resolution images of the contact geometry and the external world, as shown in Fig.~\ref{fig:visuotactile_sensor}. The key features are:

\textbf{Multimodal Perception}. By regulating the internal lighting conditions of the STS sensor, the transparency of the reflective paint coating of the sensor can be controlled, allowing the sensor to provide both visual and tactile feedback about the contacting object.

\textbf{High-Resolution Sensing}. Both visual and tactile signals are given as a high resolution image of $1640\times1232$. We use the Variable Focus Camera Module for Raspberry Pi by Odeseven, which provides a $160^{\circ}$ field of view. This results in two sensing signals that have the same point of view, frame of reference, and resolution.

\subsection{Sensor design}
\label{sec:hardware}
Inspired by recent developments in the GelSight technology \cite{yuan2017gelsight}, the STS visuotactile sensor is composed of a compliant membrane, internal illumination sources, a reflective paint layer, and a camera. When an object is pressed against the sensor, a  camera  located  within  the sensor captures the view through the ``skin'' 
as well as the deformation  of  the compliant  membrane, and produces an image that encodes tactile information, such as contact geometry, interactions forces, and stick/slip behavior. 

While optical tactile sensors typically make use of an opaque and  reflective paint coating, we developed a membrane with a controllable transparency, allowing  the sensor to provide tactile information about physical interactions and  visual  information about the world external to the sensor. This ability to
capture a visual perspective of the region beyond the contact surface enables the sensor to visualize color and the appearance of the objects as they collide with the sensor. We control the duty cycle of tactile versus visual measurements of the sensor by changing the internal
lighting condition of the STS sensor, which sets the transparency
of the reflective paint coating of the sensor. More details on the design can be found in \citet{hogan2020seeing}.

\subsection{Simulator}
\label{sec:simulator}


We developed a visuotactile simulator for the STS sensor within the PyBullet environment that reconstructs high resolution tactile signatures from the contact force and geometry. We exploit the simulator  to quickly generate large visuotactile datasets of object interactions in dynamic scenes to validate the performance of  perception models.
The simulator 
maps the geometric information of the colliding objects via the shading equation \cite{yuan2017gelsight}:
\begin{equation}
    \fontsize{8pt}{9.6pt}
    \mathbf{I}(x, y) = \mathbf{R}(\frac{\partial f}{\partial x}, \frac{\partial f}{ \partial y}),
    \label{eq:shading}
\end{equation}
where $\mathbf{I}(x, y)$ is the image intensity, $z = f(x, y)$ is the height map of the sensor surface, and $\mathbf{R}$ is the reflectance function modeling the environment lighting and surface reflectance \cite{yuan2017gelsight}. 

Following \citet{gomesgelsight}, we  implement the reflectance function $\mathbf{R}$ using Phong's reflection model, which breaks down the lighting into three main components of ambient, diffuse, and specular for each channel:
\begin{equation}
    \fontsize{8pt}{9.6pt}
    \mathbf{I}(x, y) = k_a i_a + \sum_{m \in lights} k_d (\hat{L}_{m} \cdot \hat{N}) i_{m, d} + k_s (\hat{R}_m \cdot \hat{V})^\alpha i_{m, s},
    \label{eq:phong}
\end{equation}
where $\hat{L}_m$ is the direction vector from the surface point to the light source $m$, $\hat{N}$ is the surface normal,  $\hat{R}_m$ is the reflection vector computed by $\hat{R}_m = 2 (\hat{L}_{m} \cdot \hat{N}) \hat{N} - \hat{L}_{m}$, and $\hat{V}$ is the direction vector pointing towards the camera. Additional information is provided in the supplemental material.



\section{Multimodal Perception}
\label{sec:multimodal_perception}

We present a generative multimodal perceptual system that integrates visual, tactile and 3D pose (when available) feedback within a unified framework. We make use of Multimodal Variational Autoencoders (MVAE) \cite{wu2018multimodal} to learn a shared latent representation that encodes all modalities. We further show that this embedding space can encode key information about objects such as  shape, color, and interaction forces, necessary to make inferences about intuitive physics. 


The predicted outcome of a dynamic interaction can be formulated as a self-supervision problem, where the target visual and tactile images are generated given observed context frames. Our objective is to learn a generator that maps the current available observations to the  predicted configuration of the resting state. 
We show that the  MVAE architecture can be  trained to predict the most stable and informative elements of a multimodal motion trajectory.

\subsection{Variational Autoencoders}

Generative latent variable models learn the joint distribution of the data and the unobservable representations in the form of $p_\theta(x, z) = p_\theta(z) p_\theta(x|z)$, where $p_\theta(z)$ and $p_\theta(x|z)$ denote the prior and the conditional distributions, respectively. The objective is to maximize the marginal likelihood given by $p_\theta(x) = \int p_\theta(z) p_\theta(x|z)dz$. Since the integration is in general intractable, variational autoencoders (VAE) \cite{kingma2013auto} optimize a surrogate cost, the evidence lower bound (ELBO), by approximating the  posterior $p_\theta(x|z)$ with an inference network $q_\phi(z|x)$. The ELBO loss is then given by:
\begin{equation}
    \fontsize{8pt}{9.6pt}
    \label{eq:elbo_loss}
    \text{ELBO}(x) \triangleq \: \mathbb{E}_{q_\theta(z|x)} \big[\lambda \text{log} \: p_\theta(x|z) \big] - \beta \text{KL}\big(q_\phi(z|x) || p_\theta(z) \big),
\end{equation}
where the first term denotes the reconstruction loss measuring the expectation of the likelihood of the reconstructed data given the latent variables and the second term is the Kullback-Leibler divergence between the approximate and true posterior and acts as a regularization term. In order to regularize the terms in the ELBO loss, $\beta$ \cite{higgins2016beta} and $\lambda$ are used as weights.

\subsection{Multimodal Variational Autoencoders}

The VAE uses an inference network to map the observations to a latent space, followed by a decoder to map the latent variables back to the observation space. While this approach is practical with a constant observation space, it becomes challenging when using multiple modalities, where the dimensions of the observation space vary with the availability of the modalities. For example, tactile information only becomes available when contact is made with the sensor. For such multimodal problems that present variability in the availability of data, we would require training an inference network $q(z|X)$ for each subset of modalities $X \subseteq \{x_1, x_2, \dots, x_N\}$, resulting in a total of $2^N$ combinations. To deal with this combinatorial explosion of modalities,  \citet{wu2018multimodal} propose the notion of Product of Experts (PoE) to  efficiently learn the approximate joint posterior of different modalities as the product of individual posteriors of each modality. This method has the advantage of training only $N$ inference networks, one for each modality, allowing for better scaling.

\begin{figure}[t!]
    \centering
    \includegraphics[width=.46\textwidth]{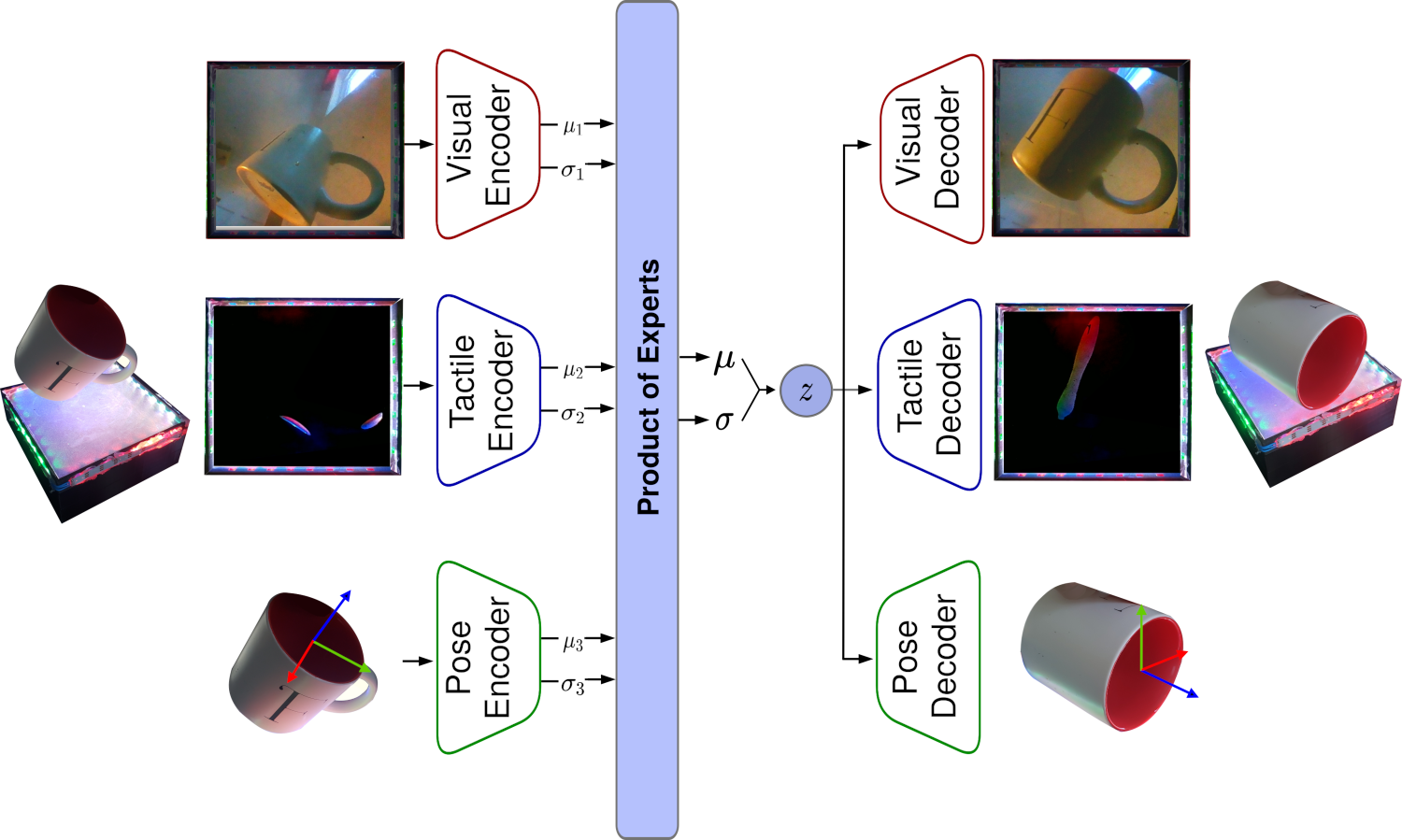}
    \caption{Multimodal dynamics modelling. A generative  perceptual system that integrates visual, tactile and 3D pose  feedback  within  a  unified Multimodal Variational Autoencoder  framework. The network gets the current  object configuration and predicts its resting configuration.}
    \label{fig:mvae}
\end{figure}

Multimodal generative modeling  learns the joint distribution of all modalities as:
\begin{equation}
    \fontsize{8pt}{9.6pt}
    p_\theta(x_1, \dots, x_N, z) = p(z) p_\theta(x_1|z) \dots p_\theta(x_N|z),
\end{equation}
where $x_i$ denotes the observation associated with mode $i$, $N$ is the total number of available modes, and $z$ is the shared latent space.
Assuming  conditional independence between modalities, we can rewrite  the joint posterior as:
\begin{align}
    \fontsize{8pt}{9.6pt}
    \label{eq:mvae_1}
    p(z|x_1, \dots, x_N) &= \frac{p(x_1, \dots, x_N|z) p(z)}{p(x_1, \dots, x_N)} \nonumber \\ 
    &= \frac{p(z)}{p(x_1, \dots, x_N)} \prod_{i=1}^{N} p(x_i|z) \nonumber \\
    &= \frac{p(z)}{p(x_1, \dots, x_N)} \prod_{i=1}^{N} \frac{p(z|x_i) p(x_i)}{p(z)} \nonumber  \\
    &\propto \frac{\prod_{i=1}^{N} p(z|x_i)}{\prod_{i=1}^{N-1} p(z)}.
\end{align}
By approximating $p(z|x_i)$ in Equation \eqref{eq:mvae_1} with $q(z|x_i) \equiv \widetilde{q}(z|x_i)p(z)$, where $\widetilde{q}(z|x_i)$ is the inference network of modality $i$, we obtain:
\begin{equation}
    \fontsize{8pt}{9.6pt}
    p(z|x_1, \dots, x_N) \propto p(z) \prod_{i=1}^{N} \widetilde{q}(z|x_i),
\end{equation}
that is recognized as the Product of Expert (PoE). In the case of variational autoencoders where $p(z)$ and $\widetilde{q}(z|x_i)$ are multivariate Gaussians, the PoE can be computed analytically as the product of two Gaussians
\cite{cao2014generalized}. 

An important advantage of the MVAE is that unlike other multimodal generative models, it can be efficiently scaled up to several modalities, as it requires training only $N$ inference models rather than the $2^N$ multimodal inference networks. Additionally, the notion of PoE allows for continuous inference in the case of discontinuous and unavailable modalities.

\subsection{Learning Intuitive Physics with MVAEs}


We use the MVAE architecture to learn a shared representation that exploits multiple sensing modalities for learning the underlying dynamics of intuitive physics. A key advantage of this formulation is that it enables combining sensing modalities, while naturally dealing with intermittent contacts, during which tactile measurements are discontinuous. 


\begin{figure}[t!]
    \centering
    \subfigure[Freefalling  objects  on  a  flat  surface.]{
    \includegraphics[width=0.45\textwidth]{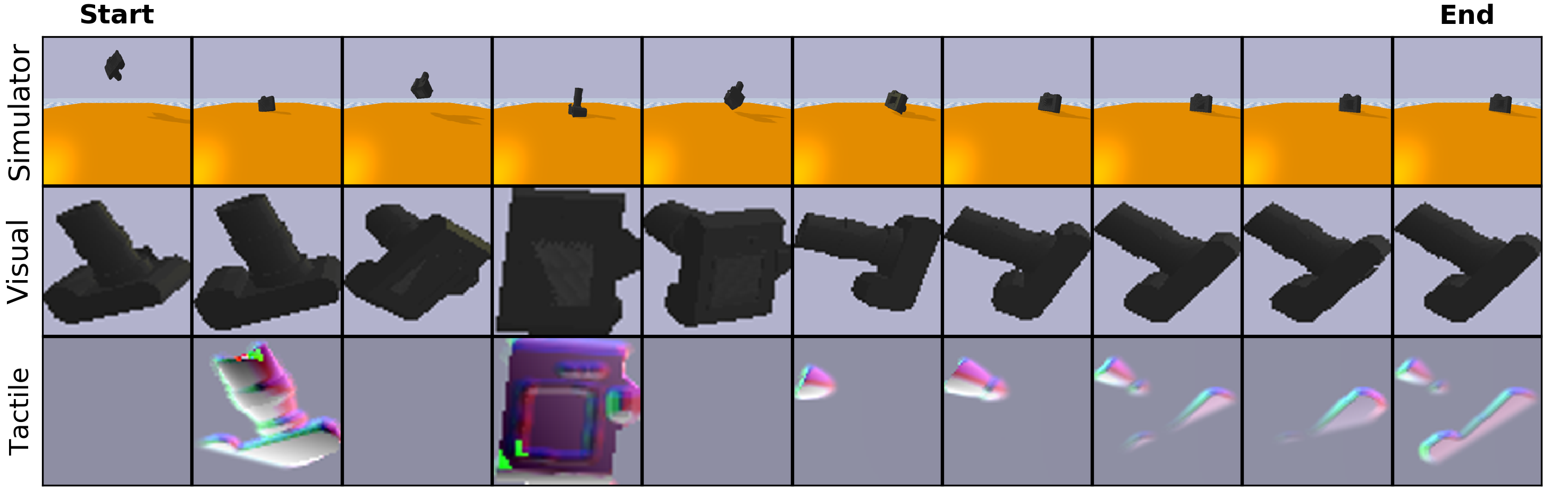}}
    \label{fig:sim_object_falling}
    \subfigure[Objects sliding down an inclined plane.]{
    \includegraphics[width=0.45\textwidth]{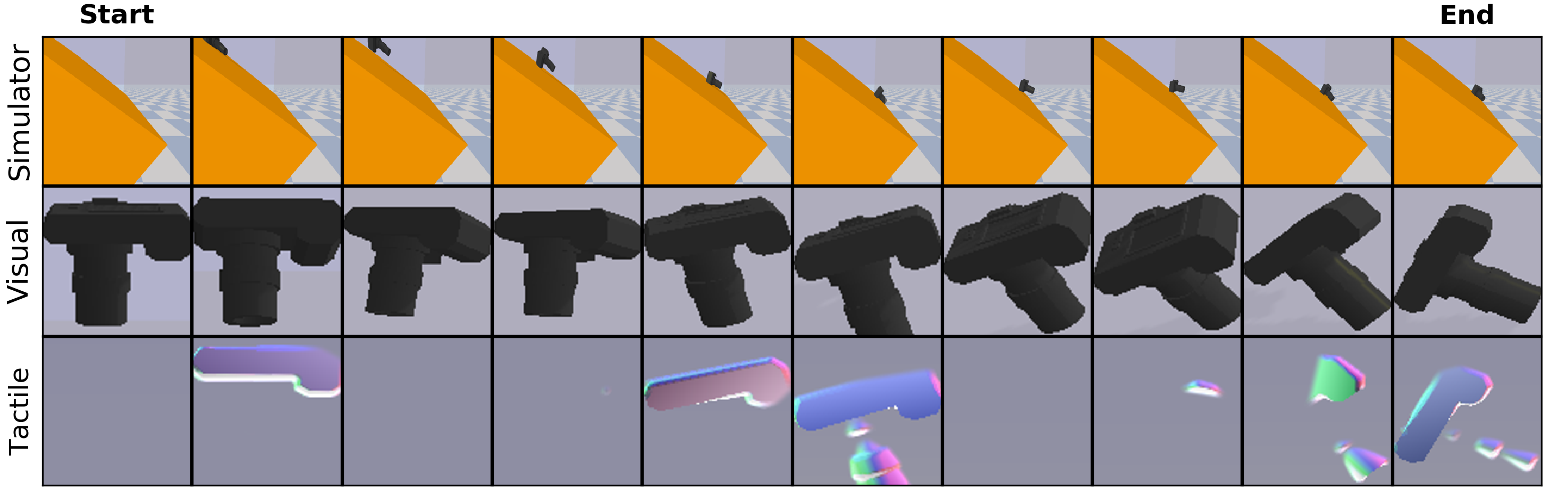}}
    \label{fig:sim_sloped}
    \subfigure[Objects perturbed from a stable resting pose.]{
    \includegraphics[width=0.45\textwidth]{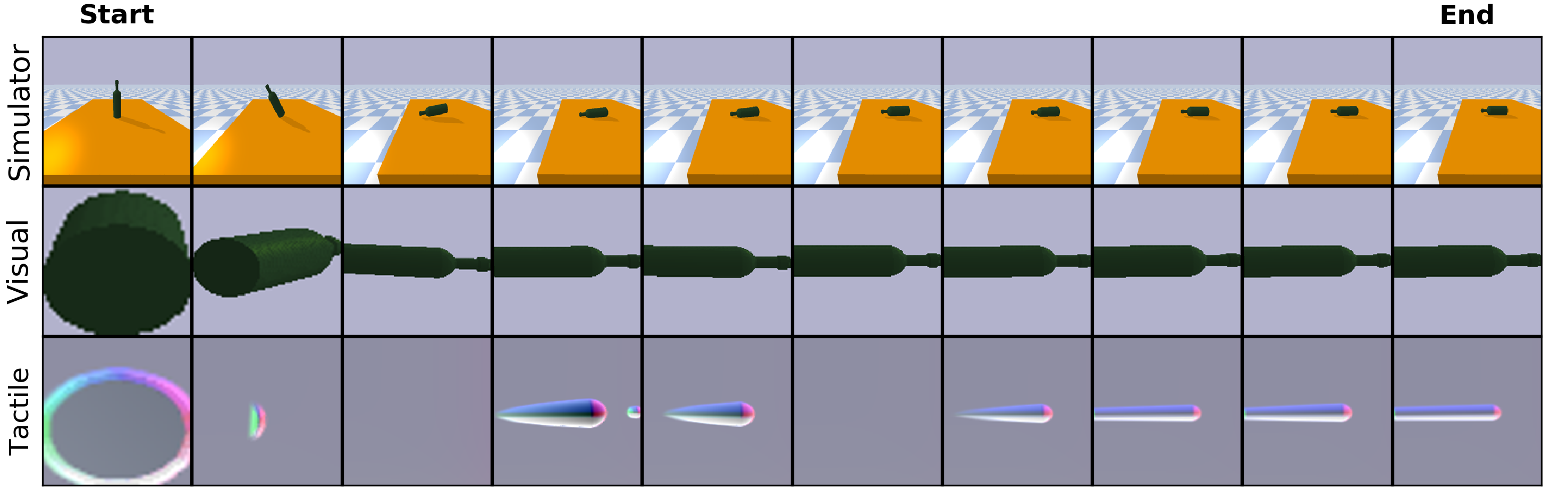}}
    \label{fig:sim_shock}
    \caption{Simulated example episodes for three dynamic simulated scenarios. The top row shows the $3$D object view while the middle and bottom rows show the visual and tactile measurement captured by the STS sensor, respectively. Some frames have been removed for clarity.}
    \label{fig:sim_tasks}
\end{figure}

While variational autoencoders are often trained by reconstructing the encoder inputs, we introduce a time-lag element into the network architecture \cite{hernandez2018variational, kipf2018neural}, where the outputs of the decoder are set to predict future frames. We adapt the ELBO loss in Eq.~\eqref{eq:elbo_loss} to:
\begin{equation}
    \fontsize{8pt}{9.6pt}
    \label{eq:dyn_elbo_loss}
    \text{ELBO}(x_t, x_T) \triangleq \: \mathbb{E}_{q_\theta(z|x_t)} \big[\lambda \text{log} \: p_\theta(x_T|z) \big] - \beta \text{KL}\big(q_\phi(z|x_t) || p_\theta(z) \big),
\end{equation}
%
%
where $t$ and $T$ denote the input and output time instances. 

Figure~\ref{fig:mvae} describes our dynamics model learning framework, where  visual, tactile, and 3D pose are fused together to learn a shared embedding space via three unimodal encoder-decoders connected through the Product of Experts. To train the model loss, we follow the sampling methodology proposed in \cite{wu2018multimodal}, where we compute the ELBO loss by enumerating the subsets of the modalities  $\mathcal{M} = \{\text{visual}, \text{tactile}, \text{pose} \}$:
\begin{equation}
\fontsize{8pt}{9.6pt}
    \mathcal{L} (x_t) = \sum_{X \in \mathbb{P}(\mathcal{M})} \text{ELBO}(X_t, X_T),
\end{equation}
where $\mathbb{P}(\mathcal{M})$ is the powerset of the modalities set $\mathcal{M}$.

In the cases where there is an input to the dynamics model (e.g., force perturbation in the third simulated scenario), we include the conditional dependence of the input condition $c$ on  the  ELBO loss as:
\begin{align}
    \fontsize{8pt}{9.6pt}
    \label{eq:dyn_elbo_loss_conditional}
    \text{ELBO}(x_t, x_T | c) \triangleq \: &\mathbb{E}_{q_\theta(z|x_t, c)} \big[\lambda \text{log} \: p_\theta(x_T|z, c) \big] \\ 
    &- \beta \text{KL}\big(q_\phi(z|x_t, c) || p_\theta(z | c) \big), \nonumber
\end{align}
%

\begin{figure}[t!]
    \centering
    \includegraphics[width=0.3\textwidth]{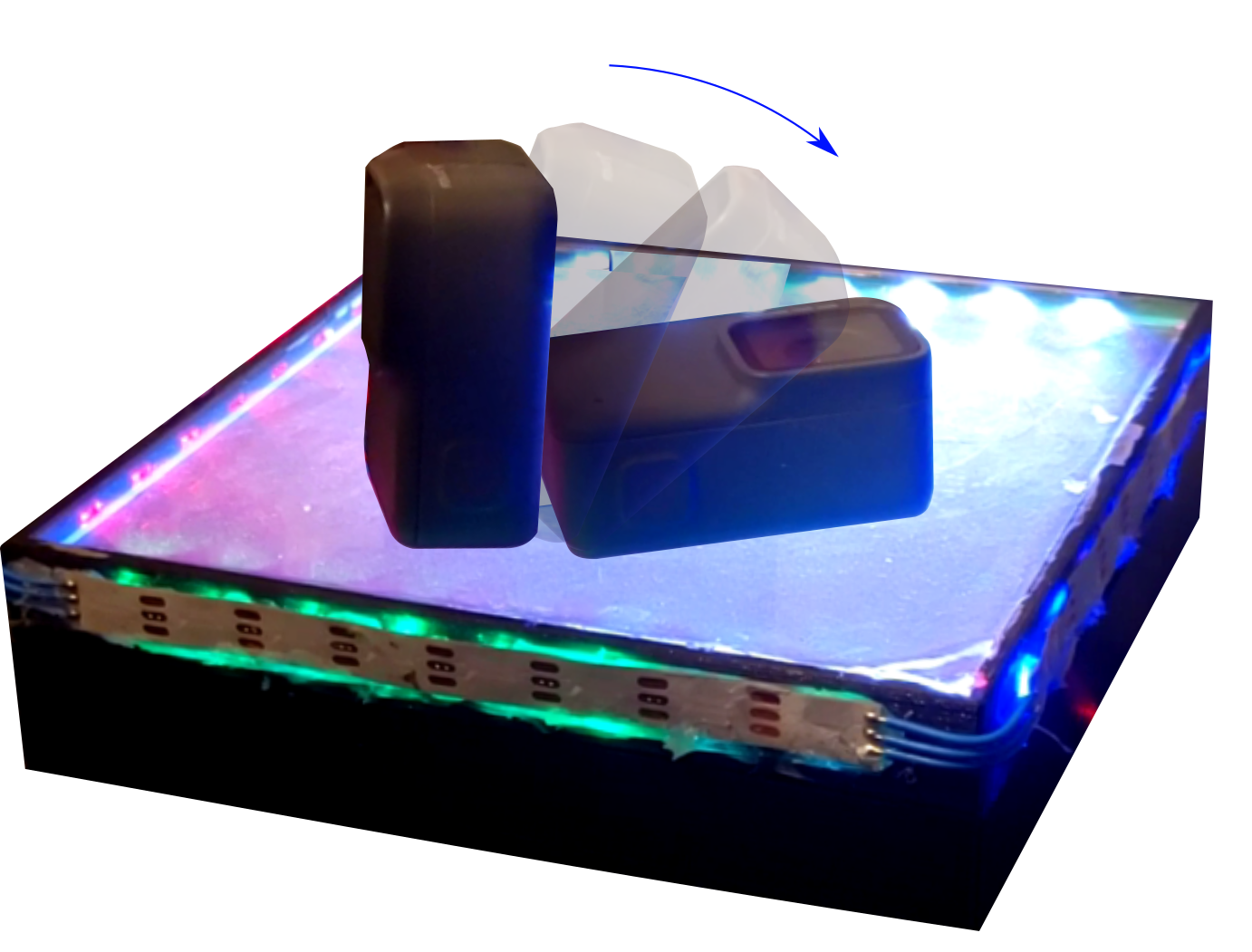}
    \caption{Real-world data collection. An electronic device (GoPro camera) is released from an unstable initial configuration. The task is to predict the resting configuration of the object from its initial measurements.}
    \label{fig:res_tasks}
\end{figure}

    
    
    

\begin{figure*}[t!]
    \centering
    \subfigure[Freefalling objects on a flat surface. The motion starts from a non-contacting initial position, accounting for the initial unavailability of  tactile measurements.]{
    \includegraphics[width=0.48\textwidth]{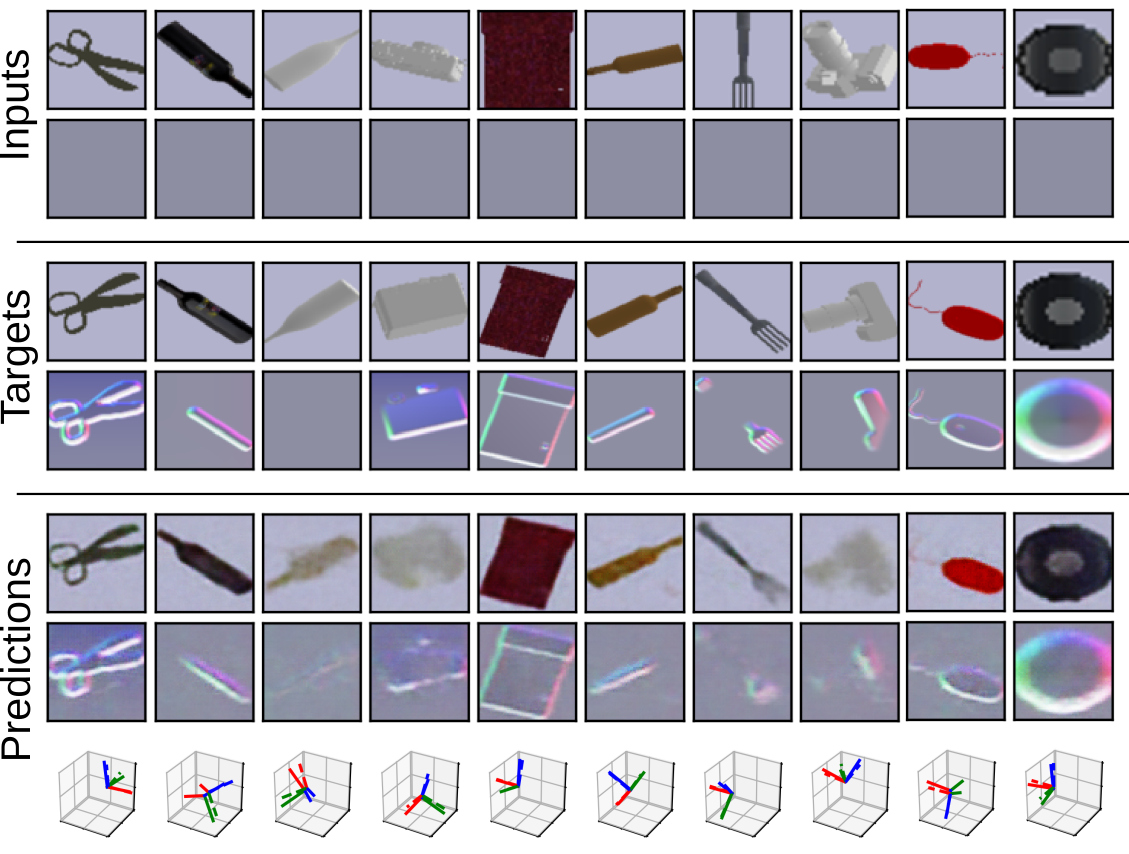}}
    \hfill
    \subfigure[Objects sliding down an inclined plane. The motion starts in proximity to the surface, accounting for the the initially availability of tactile measurements that may be either active or inactive.]{
    \includegraphics[width=0.48\textwidth]{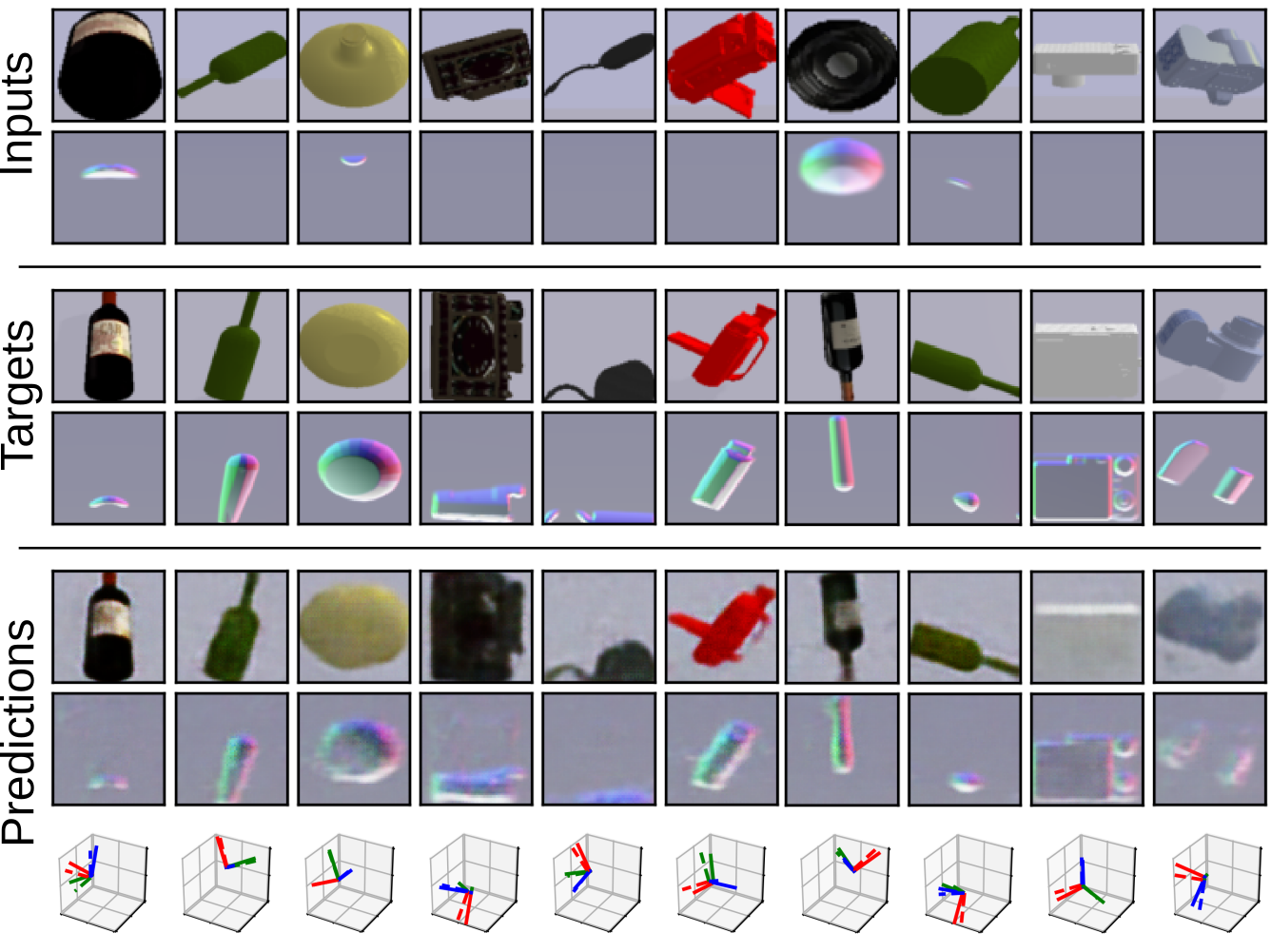}
    }
    
    
    \caption{Multimodal predictions for three simulated scenarios evaluated on the validation set. The model predicts the final resting pose in addition to the visual and tactile measurements of the STS sensor. The bottom row compares the predicted pose (solid coordinates) with the ground truth (dashed coordinates).}
    \label{fig:sim_results}
\end{figure*}

\section{Data Collection}
\label{sec:data_collection}
The simulated dataset was collected using the PyBullet simulator described earlier,
and the real-world dataset was collected using a prototype of the STS sensor.
While more details are provided in the supplemental material \footnote{Supplementary video for the experiments available at: \scriptsize \texttt{https://youtu.be/BWQ6n0mGRNc}}, we give an overview of the experimental setup.

\subsubsection{Simulated Dataset}
\label{sec:simulation_dataset}
We consider three simulated physical scenarios, as shown in Fig.~\ref{fig:sim_tasks}, involving eight household object classes\footnote{Object classes: bottle, camera, webcam, computer mouse, scissors, fork, spoon, and watch.} drawn from the 3D ShapeNet dataset \cite{chang2015shapenet}. The tasks ordered in increasing difficulty are:

\textbf{Freefalling objects on a flat surface.} This experiment releases objects with random initial poses over the STS sensor, where they collide multiple times with the sensor prior to coming to rest. We collect a total of $1700$ trajectories comprising $100$k images. 

\textbf{Objects sliding down an inclined plane.} This experiment, inspired by \citet{wu2015galileo}, places objects with random initial poses  atop an inclined surface, where they can either stick due to friction or slide down. While sliding down, the objects can roll, causing the final configuration to be significantly different than the initial. We collect a total of $2400$ trajectories comprising $145$k images. 

\textbf{Objects perturbed from a stable resting pose.} In this scenario, we consider an object initially stably resting on the sensor, that is perturbed from its equilibrium by a randomly sampled quick lateral acceleration of the sensor. This experiment only considers bottles due to their elongated and unstable shape allowing for different outcomes (e.g., toppling, sliding or standing) based on the direction and magnitude of the applied force. Due to such diverse outcomes, this task is considerably more complicated than the other two.  We collect a total of 2500 trajectories comprising $150$k images.   

\subsubsection{Real-World Dataset}
\label{sec:real_world_dataset}

We validate the predictive accuracy of our proposed framework on a small real-world dataset collected  manually using the STS sensor. We collect $2000$ images from $500$ trajectories using a small electronic device (GoPro). This object has been selected due to its small form factor (small enough to fit on the $15$cm$\times 15$cm sensor  prototype) and its mass (heavy enough to leave a meaningfully tactile signature on the sensor).   Each trajectory includes the initial and final visual and tactile images, obtained by rapidly turning on/off the internal lights of the sensor. The object is released from an unstable initial position, while being in contact with the sensor, as illustrated in Fig.~\ref{fig:res_tasks}, and the end of the episode is determined once the object is immobile.

\section{Experimental Results}
\label{ref:experimental_results}
We validate the ability of our approach to predict the evolution of physical configurations on  simulated and real scenes. We downsample the sensor images to a resolution of $64\times64$ images and an identical network architecture and training parameters  consistent across all evaluations. More details are provided in the supplemental material. 


\subsubsection{Simulation}
We compare the performance of our proposed framework against two baselines. First, we highlight the value of multimodal sensing for learning dynamics models. Second, we compare our model against dynamic models parameterized using higher temporal resolutions. 

\begin{figure}[b!]
    \centering
    \includegraphics[width=0.45\textwidth]{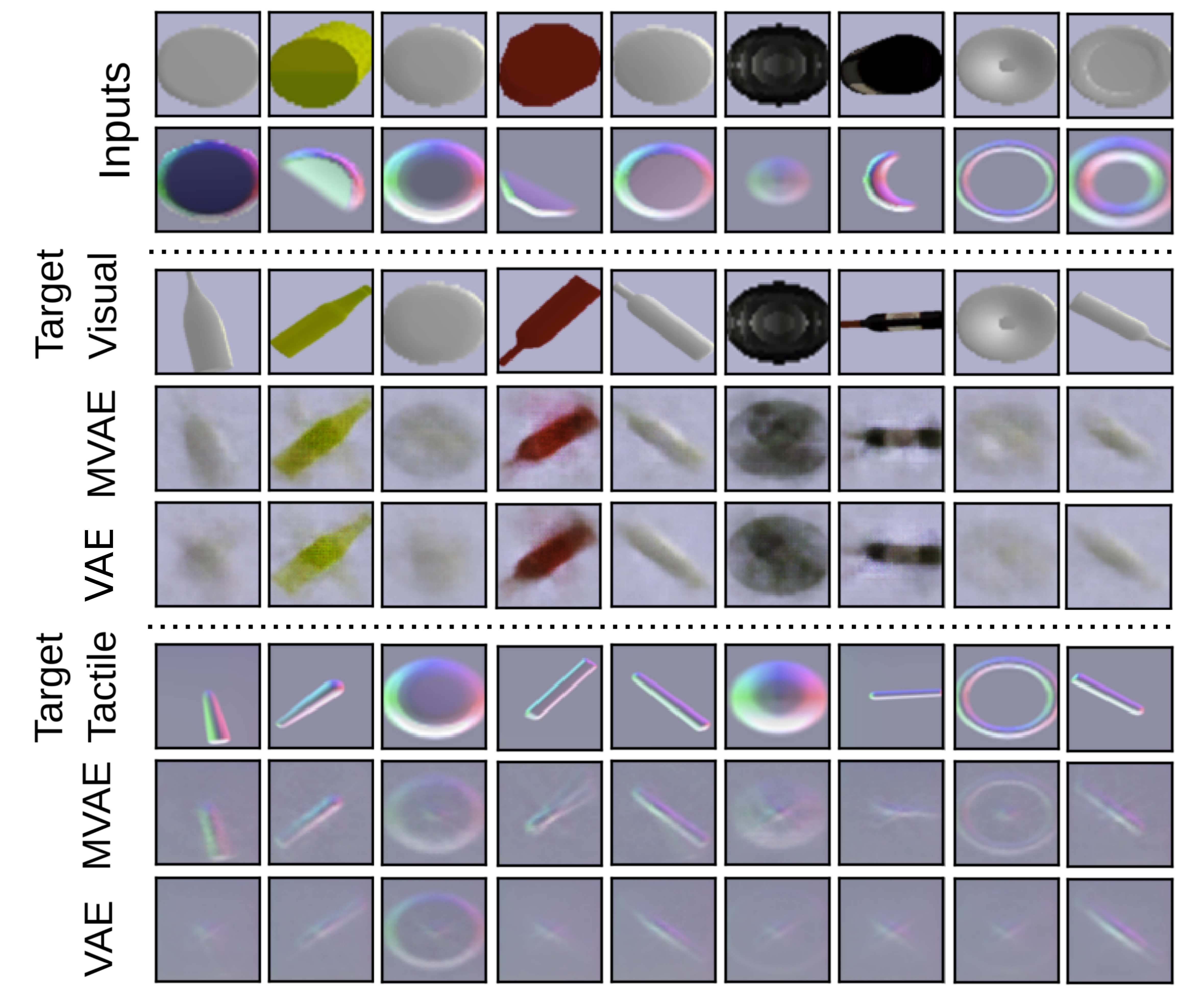}
    \caption{Qualitative comparison of visual and tactile predictions of MVAE with unimodal VAE for the simulated force perturbation scenario obtained on the validation set. MVAE leverages the tactile mode to provide clearer predictions of resting configuration.}
    \label{fig:sim_res_shock}
\end{figure}

In Fig.~\ref{fig:sim_results} and \ref{fig:sim_res_shock}, we present the  multimodal predictions for three simulated scenarios evaluated on the validation set. We show the MVAE's ability to predict  the raw visual and tactile measurements of the resting configuration of an object with high accuracy, with the predictions closely matching the ground truth labels. Interestingly, the model learns  a mapping between the visual, tactile and 3D pose modes allowing it to  sample correlating outputs from the learned shared embedding. Fig.~\ref{fig:sim_results}(a) highlights the ability of the MVAE model to handle missing modalities, such as when tactile information is unavailable in the input. Finally, the model learns to accurately predict  instances where the object has fallen from the surface of the sensor, resulting in empty output images. 

In Figure~\ref{fig:sim_res_shock}, we present the prediction results for the scenario where objects are perturbed from an initial stable pose. Unlike the first two experiments, this scenario includes a random lateral force  applied on the system that plays a significant role in determining the resting state of the objects, thus making it considerably more complicated. To account for this, we condition the MVAE with information about the magnitude and direction of the lateral force using Eq.~\eqref{eq:dyn_elbo_loss_conditional}. The results in Fig. \ref{fig:sim_res_shock} indicate that the model successfully integrates information about applied forces by correctly predicting the outcome about the object motion (i.e., toppling or falling). Further comparison with the unimodal models shows that the tactile mode has played an important role in reducing the uncertainty and blurriness of the predictions.

\begin{figure}[t!]
    \centering
    \includegraphics[width=0.45\textwidth]{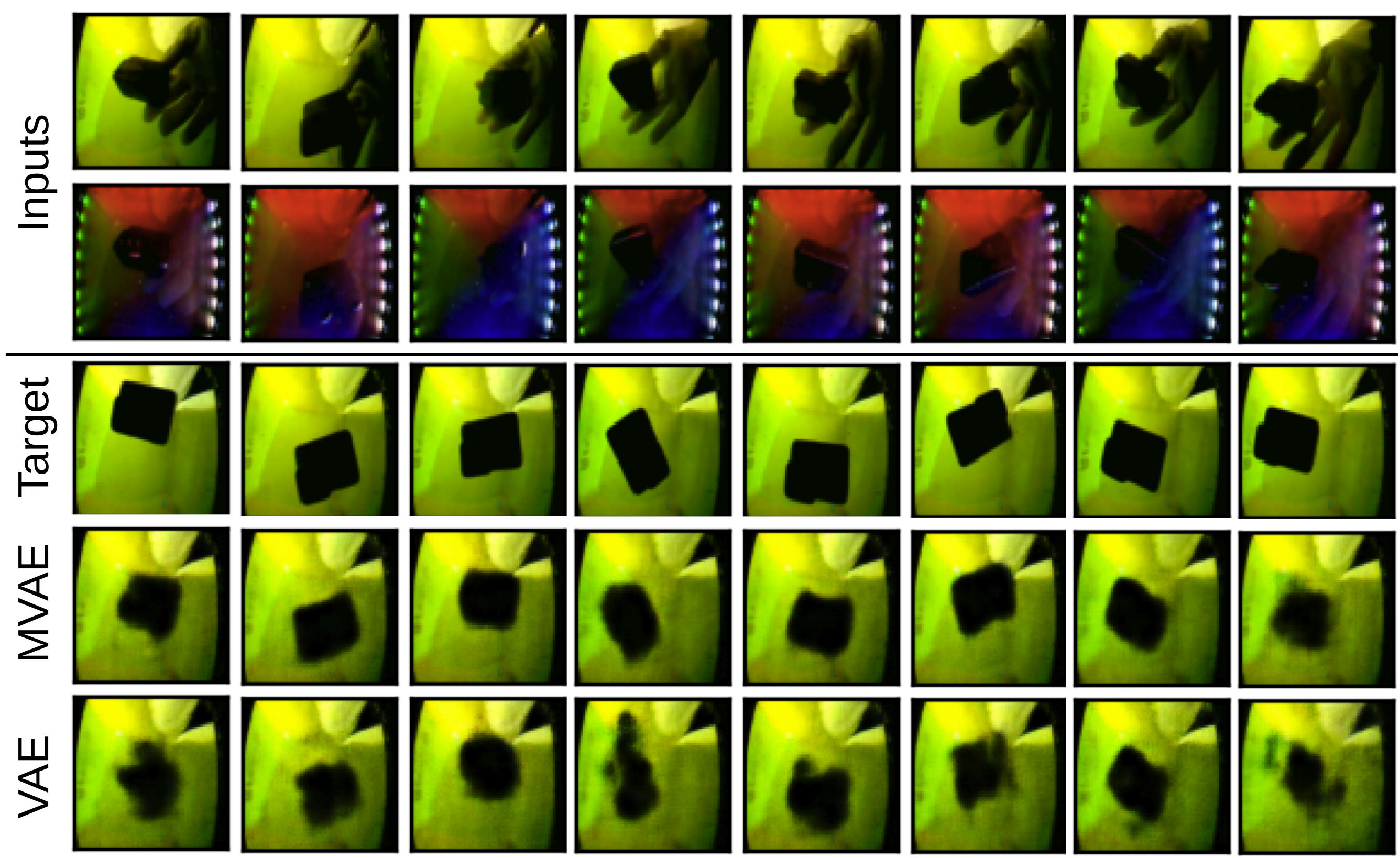}
    \caption{Qualitative comparison of visual predictions of MVAE with unimodal VAE for the real dataset obtained on the validation set. MVAE leverages the tactile mode to provide more accurate and clearer visual predictions of the resting configuration.}
    \label{fig:real_results}
\end{figure}
In Table~\ref{tab:sim_results}, we present the quantitative results comparing the multimodal and unimodal models for both one-step (high temporal resolution) and resting state predictions (resting object configuration). We  draw two conclusions from this analysis. First,  models that exploit multimodal sensing outperform  those relying on visual/tactile modalities alone. Importantly, we find that tactile information improves the prediction accuracy (for both visual and tactile predictions) by reasoning about interaction forces and the geometry of contact. Second, we show that predicting the resting state of the object outperforms dynamic models using high temporal resolution. This is due to the fact that uncertainties and errors propagate forward in time, leading to 
blurry  and  imprecise  predictions. The results show that in dynamic scenarios where the intermediate states are  not of interest, we can learn to predict the final outcome with higher accuracy without explicitly reasoning about intermediate steps. 

\begin{table}[t!]
    \centering
    \begin{tabular}{ c V{3} c| *{2}{c} | *{2}{c}}
    \hlineB{3}
    \rule{0pt}{10pt} \multirow{3}{*}{\rotatebox[origin=c]{90}{$\leftarrow$ \textbf{Task}}} & \multirow{2}{*}{\textbf{Setting} $\rightarrow$} & \multicolumn{2}{c|}{\textbf{Visual Perf.}} & \multicolumn{2}{c}{\textbf{Tactile Perf.}} \\
    \rule{0pt}{10pt} & & \multicolumn{2}{c|}{($\times$1e-4)} & \multicolumn{2}{c}{($\times$1e-4)} \\
    \rule{0pt}{10pt}& \multirow{2}{*}{\textbf{Model} $\downarrow$} & Multi & Final & Multi & Final \\
    \rule{0pt}{10pt} & & Step & Step & Step & Step \\
    \hlineB{3}
    \multirow{4}{*}{\rotatebox[origin=c]{90}{\textbf{Freefall}}} & VAE-visual only & 6522 & 5750 & NA & NA \\
    & VAE-tactile only & NA & NA & 6770  & \textbf{6703} \\
    & MVAE w/ pose & 6548  & \textbf{5741} & 6726 & \textbf{6703} \\
    & MVAE w/o pose & 6635 & 5752 & 6735 & \textbf{6703} \\
    \hlineB{3}
    \multirow{4}{*}{\rotatebox[origin=c]{90}{\textbf{Inclined}}} & VAE-visual only & 6751 & 5895 & NA & NA \\
    & VAE-tactile only & NA & NA & 6714 & 6711 \\
    & MVAE w/ pose & 6625 & \textbf{5891} & 6719 & \textbf{6709} \\
    & MVAE w/o pose & 6549 & \textbf{5890} & 6713 & \textbf{6710} \\
    \hlineB{3}
    \multirow{4}{*}{\rotatebox[origin=c]{90}{\textbf{Perturb.}}} & VAE-visual only & 7369 & 6158 & NA & NA \\
    & VAE-tactile only & NA & NA & 6879 & 6705 \\
    & MVAE w/ pose & 7552 & \textbf{6054} & 6868 & \textbf{6701} \\
    & MVAE w/o pose & 6967 & 6095 & 6896 & \textbf{6702} \\
    \hlineB{3}
\end{tabular}

    \caption{Prediction performance of fixed-step and final-step predictions for unimodal and multimodal VAE models for the three simulated scenarios. Performance is reported as the average of the binary cross-entropy error on the validation set. The bold cells indicate preferred values.}
    \label{tab:sim_results}
\end{table}



\subsubsection{Real Sensor}
This section validates the predictive abilities of the perceptual system on a real-world experiment setup using the STS sensor. 
In Figure~\ref{fig:real_results}, we show the network's ability to predict the resting object configuration, reasoning through both the visual and tactile images. Qualitative comparison of visual predictions of MVAE with unimodal VAE shows that the MVAE model leverages the tactile mode to reason more accurately about the resting configuration. In Table \ref{tab:real_results}, we quantitatively compare the predictive performances of multimodal and unimodal models,  highlighting the important role played by both visual and tactile measurements to determine the outcome of physical interactions. 

\begin{table}[t!]
    \centering
    {
\setlength{\tabcolsep}{6pt}
\begin{tabular}{ c| c |c}
    \hlineB{3}
    \textbf{Setting} $\rightarrow$ & \textbf{Visual} & \textbf{Tactile} \\
    \textbf{Model $\downarrow$} & ($\times$1e-4) & ($\times$1e-4) \\
    \hlineB{3}
    VAE-visual only & 3532 & NA \\
    VAE-tactile only & NA & 2872 \\
    MVAE w/o pose & \textbf{3509} & \textbf{2819}\\
    \hlineB{3}
    
\end{tabular}
}

    \caption{Prediction performance of final-step predictions for unimodal and multimodal VAE models on the real dataset. The performance is reported as the average of the binary cross-entropy error (BCE) on the validation set. The bold cells indicate the best values.}
    \label{tab:real_results}
\end{table}


\section{Conclusion}

We have designed and implemented a system that exploits visual and tactile feedback to make physical predictions about the motion of objects in dynamic scenes. We exploit a novel visuotactile sensor, the See-Through-your-Skin (STS) sensor, that represents both modalities as high resolution images. The perceptual system uses a multimodal variational autoencoder (MVAE) neural network architecture that maps the sensing modalities to a shared embedding, used to infer the stable resting  configuration of objects during physical interactions. By focusing on predicting the most stable elements of a trajectory, we validate the predictive abilities of our dynamics models in three simulated dynamic scenarios and a real-world experiment using the STS sensor. Results show that the MVAE framework can accurately predict the future state of objects during physical interactions with a surface. Importantly, we find that predicting object motions benefits from exploiting both modalities: visual information captures object properties such as 3D shape and location, while tactile information provides critical cues about interaction forces and resulting object motion and contacts. 

\section{Ethics Statement}
The research in this work is related to the development of smarter devices, possibly including assistive devices, but we do not foresee any other substantive ethical implications.

\bibliography{references.bib}

\begin{thebibliography}{53}
\providecommand{\natexlab}[1]{#1}
\providecommand{\url}[1]{\texttt{#1}}
\providecommand{\urlprefix}{URL }
\expandafter\ifx\csname urlstyle\endcsname\relax
  \providecommand{\doi}[1]{doi:\discretionary{}{}{}#1}\else
  \providecommand{\doi}{doi:\discretionary{}{}{}\begingroup
  \urlstyle{rm}\Url}\fi

\bibitem[{Abad and Ranasinghe(2020)}]{Abad:20}
Abad, A.; and Ranasinghe, A. 2020.
\newblock Visuotactile Sensors with Emphasis on GelSight Sensor: A Review.
\newblock \emph{IEEE Sensors Journal} PP: 1--1.
\newblock \doi{10.1109/JSEN.2020.2979662}.

\bibitem[{Allen(1984)}]{Allen1984Surface}
Allen, P. 1984.
\newblock Surface descriptions from vision and touch.
\newblock In \emph{IEEE International Conference on Robotics and Automation
  (ICRA)}, volume~1, 394--397. IEEE.

\bibitem[{Bacon, Harb, and Precup(2017)}]{bacon2017option}
Bacon, P.-L.; Harb, J.; and Precup, D. 2017.
\newblock The option-critic architecture.
\newblock In \emph{Thirty-First AAAI Conference on Artificial Intelligence}.

\bibitem[{{Bierbaum}, {Gubarev}, and {Dillmann}(2008)}]{Bierbaum2008}
{Bierbaum}, A.; {Gubarev}, I.; and {Dillmann}, R. 2008.
\newblock Robust Shape Recovery for Sparse Contact Location and Normal Data
  from Haptic Exploration.
\newblock In \emph{2008 IEEE/RSJ International Conference on Intelligent Robots
  and Systems (IROS)}, 3200--3205. Nice, France.

\bibitem[{Calandra et~al.(2018)Calandra, Owens, Jayaraman, Lin, Yuan, Malik,
  Adelson, and Levine}]{calandra2018more}
Calandra, R.; Owens, A.; Jayaraman, D.; Lin, J.; Yuan, W.; Malik, J.; Adelson,
  E.~H.; and Levine, S. 2018.
\newblock More than a feeling: Learning to grasp and regrasp using vision and
  touch.
\newblock \emph{IEEE Robotics and Automation Letters} 3(4): 3300--3307.

\bibitem[{Calandra et~al.(2017)Calandra, Owens, Upadhyaya, Yuan, Lin, Adelson,
  and Levine}]{calandra2017feeling}
Calandra, R.; Owens, A.; Upadhyaya, M.; Yuan, W.; Lin, J.; Adelson, E.~H.; and
  Levine, S. 2017.
\newblock The feeling of success: Does touch sensing help predict grasp
  outcomes?
\newblock In \emph{1st Conf. on Robot Learning (CoRL)}. Mountain View, CA.

\bibitem[{Cao and Fleet(2014)}]{cao2014generalized}
Cao, Y.; and Fleet, D.~J. 2014.
\newblock Generalized product of experts for automatic and principled fusion of
  Gaussian process predictions.
\newblock \emph{arXiv preprint arXiv:1410.7827} .

\bibitem[{Chang et~al.(2015)Chang, Funkhouser, Guibas, Hanrahan, Huang, Li,
  Savarese, Savva, Song, Su et~al.}]{chang2015shapenet}
Chang, A.~X.; Funkhouser, T.; Guibas, L.; Hanrahan, P.; Huang, Q.; Li, Z.;
  Savarese, S.; Savva, M.; Song, S.; Su, H.; et~al. 2015.
\newblock Shapenet: An information-rich 3d model repository.
\newblock \emph{arXiv preprint arXiv:1512.03012} .

\bibitem[{Donlon et~al.(2018)Donlon, Dong, Liu, Li, Adelson, and
  Rodriguez}]{donlon2018gelslim}
Donlon, E.; Dong, S.; Liu, M.; Li, J.; Adelson, E.; and Rodriguez, A. 2018.
\newblock GelSlim: A high-resolution, compact, robust, and calibrated
  tactile-sensing finger.
\newblock In \emph{2018 IEEE/RSJ International Conference on Intelligent Robots
  and Systems (IROS)}, 1927--1934. Madrid, Spain: IEEE.

\bibitem[{Driess, Englert, and Toussaint(2017)}]{17-driess-IROS}
Driess, D.; Englert, P.; and Toussaint, M. 2017.
\newblock Active Learning with Query Paths for Tactile Object Shape
  Exploration.
\newblock In \emph{IEEE/RSJ International Conference on Intelligent Robots and
  Systems (IROS)}. Vancouver, Canada.

\bibitem[{Fazeli et~al.(2020)Fazeli, Zapolsky, Drumwright, and
  Rodriguez}]{fazeli2020fundamental}
Fazeli, N.; Zapolsky, S.; Drumwright, E.; and Rodriguez, A. 2020.
\newblock Fundamental limitations in performance and interpretability of common
  planar rigid-body contact models.
\newblock In Amato, N.; Hager, G.; Thomas, S.; and Torres-Torriti, M., eds.,
  \emph{Robotics Research}, 555--571. Springer.

\bibitem[{Gomes, Wilson, and Luo(2019)}]{gomesgelsight}
Gomes, D.~F.; Wilson, A.; and Luo, S. 2019.
\newblock GelSight Simulation for Sim2Real Learning.
\newblock In \emph{CRA ViTac Workshop}. Montreal, Canada.
\newblock Held in conjunction with IEEE ICRA.

\bibitem[{Hern{\'a}ndez et~al.(2018)Hern{\'a}ndez, Wayment-Steele, Sultan,
  Husic, and Pande}]{hernandez2018variational}
Hern{\'a}ndez, C.~X.; Wayment-Steele, H.~K.; Sultan, M.~M.; Husic, B.~E.; and
  Pande, V.~S. 2018.
\newblock Variational encoding of complex dynamics.
\newblock \emph{Physical Review E} 97(6): 062412.

\bibitem[{Higgins et~al.(2017)Higgins, Matthey, Pal, Burgess, Glorot,
  Botvinick, Mohamed, and Lerchner}]{higgins2016beta}
Higgins, I.; Matthey, L.; Pal, A.; Burgess, C.; Glorot, X.; Botvinick, M.;
  Mohamed, S.; and Lerchner, A. 2017.
\newblock beta-vae: Learning basic visual concepts with a constrained
  variational framework.
\newblock In \emph{International Conference on Learning Representations
  (ICLR)}. Toulon, France.

\bibitem[{Hogan et~al.(2020)Hogan, Jenkin, Rezaei-Shoshtari, Girdhar, Meger,
  and Dudek}]{hogan2020seeing}
Hogan, F.~R.; Jenkin, M.; Rezaei-Shoshtari, S.; Girdhar, Y.; Meger, D.; and
  Dudek, G. 2020.
\newblock Seeing Through your Skin: Recognizing Objects with a Novel
  Visuotactile Sensor.
\newblock \emph{arXiv preprint arXiv:2011.09552} .

\bibitem[{Izatt et~al.(2017)Izatt, Mirano, Adelson, and
  Tedrake}]{izatt2017tracking}
Izatt, G.; Mirano, G.; Adelson, E.; and Tedrake, R. 2017.
\newblock Tracking objects with point clouds from vision and touch.
\newblock In \emph{2017 IEEE International Conference on Robotics and
  Automation (ICRA)}, 4000--4007. IEEE.

\bibitem[{Jayaraman et~al.(2018)Jayaraman, Ebert, Efros, and
  Levine}]{jayaraman2018time}
Jayaraman, D.; Ebert, F.; Efros, A.; and Levine, S. 2018.
\newblock Time-Agnostic Prediction: Predicting Predictable Video Frames.
\newblock In \emph{International Conference on Learning Representations
  (ICLR)}. Vancouver, Canada.

\bibitem[{Johnson and Adelson(2009)}]{Johnson:09}
Johnson, M.~K.; and Adelson, E. 2009.
\newblock Retrographic sensing for the measurement of surface texture and
  shape.
\newblock In \emph{IEEE Int. Conf. on Computer Vision and Pattern Recognition
  (CVPR)}, 1070--1077. Miami, FL.

\bibitem[{Kingma and Ba(2014)}]{kingma2014adam}
Kingma, D.~P.; and Ba, J. 2014.
\newblock Adam: A method for stochastic optimization.
\newblock \emph{arXiv preprint arXiv:1412.6980} .

\bibitem[{Kingma and Welling(2013)}]{kingma2013auto}
Kingma, D.~P.; and Welling, M. 2013.
\newblock Auto-encoding variational bayes.
\newblock \emph{arXiv preprint arXiv:1312.6114} .

\bibitem[{Kipf et~al.(2018)Kipf, Fetaya, Wang, Welling, and
  Zemel}]{kipf2018neural}
Kipf, T.; Fetaya, E.; Wang, K.-C.; Welling, M.; and Zemel, R. 2018.
\newblock Neural Relational Inference for Interacting Systems.
\newblock In \emph{International Conference on Machine Learning}, 2688--2697.

\bibitem[{Kuppuswamy et~al.(2020)Kuppuswamy, Alspach, Uttamchandani, Creasey,
  Ikeda, and Tedrake}]{kuppuswamy2020soft}
Kuppuswamy, N.; Alspach, A.; Uttamchandani, A.; Creasey, S.; Ikeda, T.; and
  Tedrake, R. 2020.
\newblock Soft-Bubble grippers for robust and perceptive manipulation.
\newblock \emph{arXiv preprint arXiv:2004.03691} .

\bibitem[{Lambeta et~al.(2020)Lambeta, Chou, Tian, Yang, Maloon, Most, Stroud,
  Santos, Byagowi, Kammerer et~al.}]{lambeta2020digit}
Lambeta, M.; Chou, P.-W.; Tian, S.; Yang, B.; Maloon, B.; Most, V.~R.; Stroud,
  D.; Santos, R.; Byagowi, A.; Kammerer, G.; et~al. 2020.
\newblock DIGIT: A Novel Design for a Low-Cost Compact High-Resolution Tactile
  Sensor With Application to In-Hand Manipulation.
\newblock \emph{IEEE Robotics and Automation Letters} 5(3): 3838--3845.

\bibitem[{Lee, Bollegala, and Luo(2019)}]{lee2019touching}
Lee, J.-T.; Bollegala, D.; and Luo, S. 2019.
\newblock “Touching to See” and “Seeing to Feel”: Robotic Cross-modal
  Sensory Data Generation for Visual-Tactile Perception.
\newblock In \emph{2019 International Conference on Robotics and Automation
  (ICRA)}, 4276--4282. IEEE.

\bibitem[{{Lee} et~al.(2019){Lee}, {Zhu}, {Srinivasan}, {Shah}, {Savarese},
  {Fei-Fei}, {Garg}, and {Bohg}}]{Lee2019}
{Lee}, M.~A.; {Zhu}, Y.; {Srinivasan}, K.; {Shah}, P.; {Savarese}, S.;
  {Fei-Fei}, L.; {Garg}, A.; and {Bohg}, J. 2019.
\newblock Making Sense of Vision and Touch: Self-Supervised Learning of
  Multimodal Representations for Contact-Rich Tasks.
\newblock In \emph{2019 International Conference on Robotics and Automation
  (ICRA)}, 8943--8950.

\bibitem[{Lee et~al.(2019)Lee, Zhu, Srinivasan, Shah, Savarese, Fei-Fei, Garg,
  and Bohg}]{lee2019making}
Lee, M.~A.; Zhu, Y.; Srinivasan, K.; Shah, P.; Savarese, S.; Fei-Fei, L.; Garg,
  A.; and Bohg, J. 2019.
\newblock Making sense of vision and touch: Self-supervised learning of
  multimodal representations for contact-rich tasks.
\newblock In \emph{2019 International Conference on Robotics and Automation
  (ICRA)}, 8943--8950. IEEE.

\bibitem[{Li et~al.(2014)Li, Platt, Yuan, ten Pas, Roscup, Srinivasan, and
  Adelson}]{li2014localization}
Li, R.; Platt, R.; Yuan, W.; ten Pas, A.; Roscup, N.; Srinivasan, M.~A.; and
  Adelson, E. 2014.
\newblock Localization and manipulation of small parts using gelsight tactile
  sensing.
\newblock In \emph{IEEE/RSJ Int. Conf. on Intelligent Robots and Systems},
  3988--3993. Hong Kong, China: IEEE.

\bibitem[{Li et~al.(2019)Li, Zhu, Tedrake, and Torralba}]{li2019connecting}
Li, Y.; Zhu, J.-Y.; Tedrake, R.; and Torralba, A. 2019.
\newblock Connecting touch and vision via cross-modal prediction.
\newblock In \emph{IEEE Conf. on Computer Vision and Pattern Recognition
  (CVPR)}, 10609--10618. Long Beach, CA.

\bibitem[{Luo et~al.(2016)Luo, W.~Mou, Althoefer, and Liu}]{Luo2016}
Luo, S.; W.~Mou, W.; Althoefer, K.; and Liu, H. 2016.
\newblock Iterative Closest Labeled Point for Tactile Object Shape Recognition.
\newblock In \emph{IEEE/RSJ IROS}, 3137--3142. Daejon, Korea.
\newblock \doi{10.1109/IROS.2016.7759485}.

\bibitem[{Luo et~al.(2018)Luo, Yuan, Adelson, Cohn, and Fuentes}]{luo2018vitac}
Luo, S.; Yuan, W.; Adelson, E.; Cohn, A.~G.; and Fuentes, R. 2018.
\newblock Vitac: Feature sharing between vision and tactile sensing for cloth
  texture recognition.
\newblock In \emph{2018 IEEE International Conference on Robotics and
  Automation (ICRA)}, 2722--2727. IEEE.

\bibitem[{McGovern and Barto(2001)}]{mcgovern2001automatic}
McGovern, A.; and Barto, A.~G. 2001.
\newblock Automatic Discovery of Subgoals in Reinforcement Learning using
  Diverse Density.
\newblock In \emph{Proceedings of the Eighteenth International Conference on
  Machine Learning}, 361--368.

\bibitem[{Nair and Finn(2019)}]{nair2019hierarchical}
Nair, S.; and Finn, C. 2019.
\newblock Hierarchical Foresight: Self-Supervised Learning of Long-Horizon
  Tasks via Visual Subgoal Generation.
\newblock In \emph{International Conference on Learning Representations}.

\bibitem[{Nair, Savarese, and Finn(2020)}]{nair2020goal}
Nair, S.; Savarese, S.; and Finn, C. 2020.
\newblock Goal-Aware Prediction: Learning to Model What Matters.
\newblock \emph{arXiv preprint arXiv:2007.07170} .

\bibitem[{Neitz et~al.(2018)Neitz, Parascandolo, Bauer, and
  Sch{\"o}lkopf}]{neitz2018adaptive}
Neitz, A.; Parascandolo, G.; Bauer, S.; and Sch{\"o}lkopf, B. 2018.
\newblock Adaptive skip intervals: Temporal abstraction for recurrent dynamical
  models.
\newblock In \emph{Advances in Neural Information Processing Systems},
  9816--9826.

\bibitem[{Ottenhaus et~al.(2016)Ottenhaus, Miller, Schiebener, Vahrenkamp, and
  Asfour}]{Ottenhaus2016}
Ottenhaus, S.; Miller, M.; Schiebener, D.; Vahrenkamp, N.; and Asfour, T. 2016.
\newblock Local implicit surface estimation for haptic exploration.
\newblock In \emph{IEEE-RAS International Conference on Humanoid Robots
  (Humanoids)}, 850--856.
\newblock \doi{10.1109/HUMANOIDS.2016.7803372}.

\bibitem[{Padmanabha et~al.(2020)Padmanabha, Ebert, Tian, Calandra, Finn, and
  Levine}]{padmanabha2020omnitact}
Padmanabha, A.; Ebert, F.; Tian, S.; Calandra, R.; Finn, C.; and Levine, S.
  2020.
\newblock OmniTact: A Multi-Directional High Resolution Touch Sensor.

\bibitem[{Paszke et~al.(2017)Paszke, Gross, Chintala, Chanan, Yang, DeVito,
  Lin, Desmaison, Antiga, and Lerer}]{paszke2017automatic}
Paszke, A.; Gross, S.; Chintala, S.; Chanan, G.; Yang, E.; DeVito, Z.; Lin, Z.;
  Desmaison, A.; Antiga, L.; and Lerer, A. 2017.
\newblock Automatic differentiation in pytorch .

\bibitem[{Pertsch et~al.(2020)Pertsch, Rybkin, Yang, Zhou, Derpanis,
  Daniilidis, Lim, and Jaegle}]{pertsch2020keyframing}
Pertsch, K.; Rybkin, O.; Yang, J.; Zhou, S.; Derpanis, K.; Daniilidis, K.; Lim,
  J.; and Jaegle, A. 2020.
\newblock Keyframing the Future: Keyframe Discovery for Visual Prediction and
  Planning.
\newblock In \emph{Learning for Dynamics and Control}, 969--979. PMLR.

\bibitem[{Radford, Metz, and Chintala(2015)}]{radford2015unsupervised}
Radford, A.; Metz, L.; and Chintala, S. 2015.
\newblock Unsupervised representation learning with deep convolutional
  generative adversarial networks.
\newblock \emph{arXiv preprint arXiv:1511.06434} .

\bibitem[{Ramachandran, Zoph, and Le(2017)}]{ramachandran2017searching}
Ramachandran, P.; Zoph, B.; and Le, Q.~V. 2017.
\newblock Searching for activation functions.
\newblock \emph{arXiv preprint arXiv:1710.05941} .

\bibitem[{Shimonomura(2019)}]{shimonomura2019tactile}
Shimonomura, K. 2019.
\newblock Tactile image sensors employing camera: A review.
\newblock \emph{Sensors} 19(18): 3933.

\bibitem[{Smith et~al.(2020)Smith, Calandra, Romero, Gkioxari, Meger, Malik,
  and Drozdzal}]{smith20203d}
Smith, E.; Calandra, R.; Romero, A.; Gkioxari, G.; Meger, D.; Malik, J.; and
  Drozdzal, M. 2020.
\newblock 3d shape reconstruction from vision and touch.
\newblock \emph{Advances in Neural Information Processing Systems} 33.

\bibitem[{Sutton, Precup, and Singh(1999)}]{sutton1999between}
Sutton, R.~S.; Precup, D.; and Singh, S. 1999.
\newblock Between MDPs and semi-MDPs: A framework for temporal abstraction in
  reinforcement learning.
\newblock \emph{Artificial intelligence} 112(1-2): 181--211.

\bibitem[{Tremblay et~al.(2020)Tremblay, Manderson, Noca, Dudek, and
  Meger}]{tremblay2020multimodal}
Tremblay, J.-F.; Manderson, T.; Noca, A.; Dudek, G.; and Meger, D. 2020.
\newblock Multimodal dynamics modeling for off-road autonomous vehicles.
\newblock \emph{arXiv preprint arXiv:2011.11751} .

\bibitem[{Vlack et~al.(2005)Vlack, Mizota, Kawakami, Kamiyama, Kajimoto, and
  Tachi}]{vlack2005gelforce}
Vlack, K.; Mizota, T.; Kawakami, N.; Kamiyama, K.; Kajimoto, H.; and Tachi, S.
  2005.
\newblock GelForce: a vision-based traction field computer interface.
\newblock In \emph{CHI'05 extended abstracts on Human factors in computing
  systems}, 1154--1155.

\bibitem[{Wang et~al.(2018)Wang, Wu, Sun, Yuan, Freeman, Tenenbaum, and
  Adelson}]{wang20183d}
Wang, S.; Wu, J.; Sun, X.; Yuan, W.; Freeman, W.~T.; Tenenbaum, J.~B.; and
  Adelson, E.~H. 2018.
\newblock 3d shape perception from monocular vision, touch, and shape priors.
\newblock In \emph{2018 IEEE/RSJ International Conference on Intelligent Robots
  and Systems (IROS)}, 1606--1613. IEEE.

\bibitem[{Watkins-Valls, Varley, and Allen(2019)}]{watkins2019multi}
Watkins-Valls, D.; Varley, J.; and Allen, P. 2019.
\newblock Multi-modal geometric learning for grasping and manipulation.
\newblock In \emph{2019 International conference on robotics and automation
  (ICRA)}, 7339--7345. IEEE.

\bibitem[{Wu et~al.(2015)Wu, Yildirim, Lim, Freeman, and
  Tenenbaum}]{wu2015galileo}
Wu, J.; Yildirim, I.; Lim, J.~J.; Freeman, B.; and Tenenbaum, J. 2015.
\newblock Galileo: Perceiving physical object properties by integrating a
  physics engine with deep learning.
\newblock In \emph{Advances in neural information processing systems},
  127--135.

\bibitem[{Wu and Goodman(2018)}]{wu2018multimodal}
Wu, M.; and Goodman, N. 2018.
\newblock Multimodal generative models for scalable weakly-supervised learning.
\newblock In \emph{32nd Conf. on Neural Information Processing Systems
  (NeurIPS)}, 5575--5585. Montreal, Canada.

\bibitem[{Yamaguchi and Atkeson(2017)}]{yamaguchi2017implementing}
Yamaguchi, A.; and Atkeson, C.~G. 2017.
\newblock Implementing tactile behaviors using fingervision.
\newblock In \emph{2017 IEEE-RAS 17th International Conference on Humanoid
  Robotics (Humanoids)}, 241--248. IEEE.

\bibitem[{Yi et~al.(2016)Yi, Calandra, Filipe, {van Hoof}, Tucker, Yilei, and
  Peters}]{Yi2016Active}
Yi, Z.; Calandra, R.; Filipe, F.~V.; {van Hoof}, H.; Tucker, H.; Yilei, Z.; and
  Peters, J. 2016.
\newblock Active Tactile Object Exploration with {Gaussian} Processes.
\newblock In \emph{IEEE/RSJ International Conference on Intelligent Robots and
  Systems (IROS)}, 4925--4930. Stockholm, Sweden.
\newblock \doi{10.1109/IROS.2016.7759723}.

\bibitem[{Yuan, Dong, and Adelson(2017)}]{yuan2017gelsight}
Yuan, W.; Dong, S.; and Adelson, E.~H. 2017.
\newblock Gelsight: High-resolution robot tactile sensors for estimating
  geometry and force.
\newblock \emph{Sensors} 17(12): 2762.

\bibitem[{Yuan et~al.(2017)Yuan, Wang, Dong, and Adelson}]{yuan2017connecting}
Yuan, W.; Wang, S.; Dong, S.; and Adelson, E. 2017.
\newblock Connecting look and feel: Associating the visual and tactile
  properties of physical materials.
\newblock In \emph{Proceedings of the IEEE Conference on Computer Vision and
  Pattern Recognition}, 5580--5588.

\end{thebibliography}

\pagebreak
\begin{center}
    {\LARGE \bf Supplemental Material}    
\end{center}

\section{Implementation Details}
In this section, we present the implementation details for our methods. We implemented our models in PyTorch \cite{paszke2017automatic} and trained them on Tesla M40 and K80 GPUs.

\subsection{Network Architecture}
All evaluations use an identical network architecture.

We use a DCGAN-based \cite{radford2015unsupervised} architecture for the image-based encoder-decoders. The image encoder uses four blocks of 4$\times$4 convolution-batchnorm-Swish \cite{ramachandran2017searching} to transform input images as 3$\times$64$\times$64 $\rightarrow$ 32$\times$32$\times$32 $\rightarrow$ 64$\times$16$\times$16 $\rightarrow$ 128$\times$8$\times$8 $\rightarrow$ 256$\times$5$\times$5. A two-layer fully connected network maps the CNN features to a 256 dimensional latent space. The pose encoder uses a two-layer fully connected network to map the input poses to the latent space. Whenever the conditional VAE or MVAE was used, the condition vector has been concatenated to the latent vector at the bottleneck. 

The image decoder first upsamples the 256 dimensional latent vector through a fully-connected layer with Swish activation function and applies four blocks of 4$\times$4 transposed convolution-batchnorm-Swish to transform the input latent code as 256$\times$5$\times$5 $\rightarrow$ 128$\times$8$\times$8 $\rightarrow$ 64$\times$16$\times$16 $\rightarrow$ 32$\times$32$\times$32 $\rightarrow$ 3$\times$64$\times$64. The final block uses a linear activation function rather than the standard Sigmoid, and instead we used binary cross-entropy (BCE) loss with logits to improve the numerical stability during training. The pose decoder uses a two-layer fully connected network to map the latent code to the pose.

\subsection{Training Details}
We trained our models with the Adam optimizer \cite{kingma2014adam} for 200 epochs with batch size 64. In the ELBO loss, eq. 3 of the paper, we annealed $\beta$ from 0 to 1 during the first 50 epochs and used $\lambda=1$ for visual and tactile modes and $\lambda=1000$ for the pose mode. In all cases, $80 \%$ of the dataset has been used for training and the rest for validation. For the simulation results, to let the network better focus on the object itself, we crop visual and tactile images to the segmentation mask of the object. However, for the real results, the raw output of the actual sensor is used to train the model. 

\section{Simulator Details}
The visuotactile simulator, developed based on PyBullet, reconstructs high resolution tactile signatures from the contact form and geometry via the shading equation \cite{yuan2017gelsight}:
\begin{equation}
    \fontsize{8pt}{9.6pt}
    \mathbf{I}(x, y) = \mathbf{R}(\frac{\partial f}{\partial x}, \frac{\partial f}{ \partial y}),
    \nonumber
\end{equation}
where $\mathbf{I}(x, y)$ is the image intensity, $z = f(x, y)$ is the height map of the sensor surface, and $\mathbf{R}$ is the reflectance function modeling the environment lighting and surface reflectance \cite{yuan2017gelsight}. The surface function $f$ is  obtained  from the depth buffer provided by  OpenGL camera in PyBullet, which we clip to the thickness of the STS elastomer ($5$mm). To compute the surface normal at each point, we locate its adjacent points and calculate their principal axis using covariance analysis.

We  implement the reflectance function $\mathbf{R}$ using Phong's reflection model, which breaks down the lighting into three main components of ambient, diffuse, and specular for each channel:
\begin{equation}
    \fontsize{8pt}{9.6pt}
    \mathbf{I}(x, y) = k_a i_a + \sum_{m \in lights} k_d (\hat{L}_{m} \cdot \hat{N}) i_{m, d} + k_s (\hat{R}_m \cdot \hat{V})^\alpha i_{m, s},
    \nonumber
\end{equation}
where $\hat{L}_m$ is the direction vector from the surface point to the light source $m$, $\hat{N}$ is the surface normal,  $\hat{R}_m$ is the reflection vector computed by $\hat{R}_m = 2 (\hat{L}_{m} \cdot \hat{N}) \hat{N} - \hat{L}_{m}$, and $\hat{V}$ is the direction vector pointing towards the camera. 

Through extensive search and taking into account the suggested parameters in \citet{gomesgelsight}, we set the specular reflection constant $k_s$ to 0.5, the diffuse reflection constant $k_d$ to 1.0, the ambient reflection constant $k_a$ to 0.8, the shininess constant $\alpha$ to 5, and the RGB channels of specular and diffuse intensities ($i_s$ and $i_d$) of each corresponding light source to 1.0. We then apply a darkening mask based on the pixel penetration depth.

\begin{figure}[t!]
    \centering
    \includegraphics[width=0.45\textwidth]{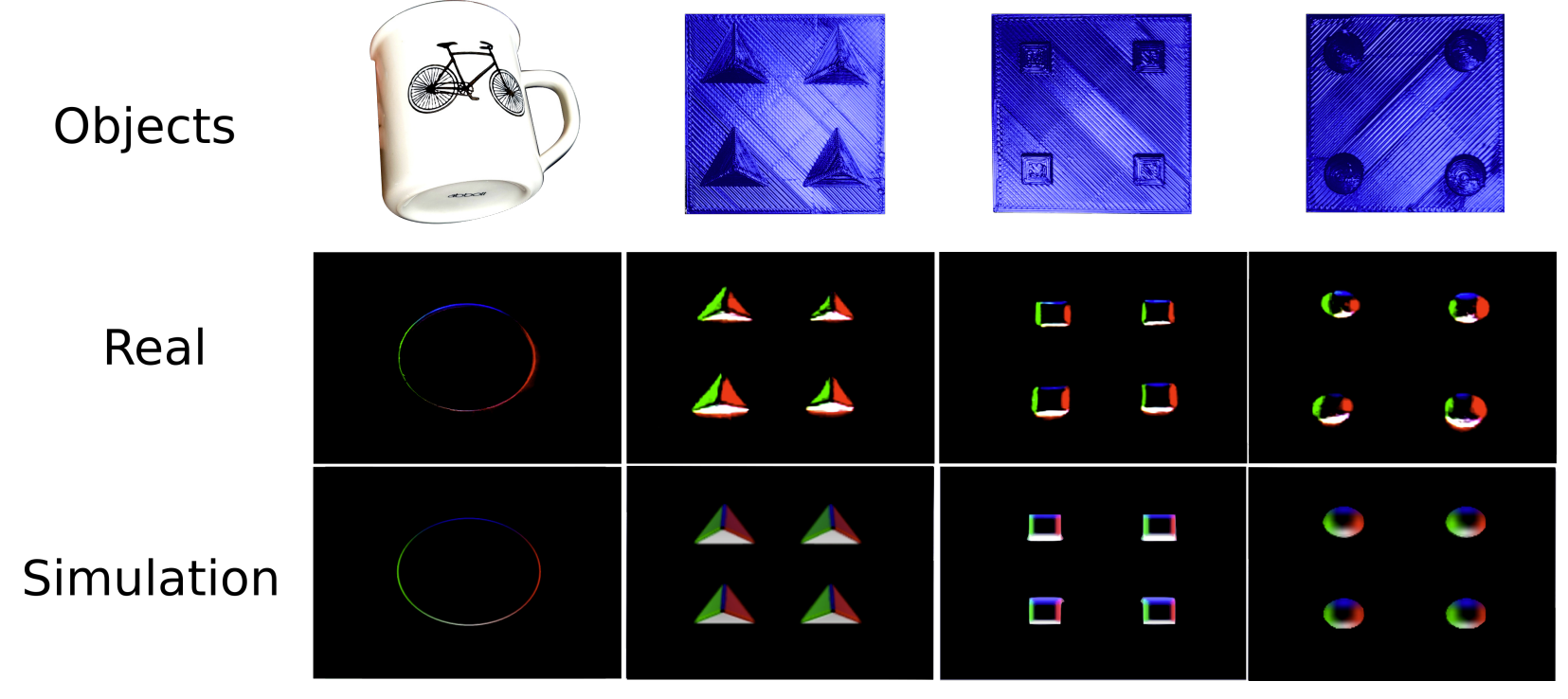}
    \caption{Comparison of simulated and real tactile imprints of similar objects for the STS sensor. Image taken from \citet{hogan2020seeing} with permission from the authors.}
    \label{fig:real_vs_sim}
\end{figure}

In order to account for the weight of objects, we introduce a lumped-element compliance model, which approximates the continuous deformations of the elastomer by an array of springs (one per pixel). Given the applied contact forces and geometries provided by the PyBullet simulator, we solve for static equilibrium at each time step to determine the gel's deformability. Although there exists a reality gap between the simulator and the real sensor, mostly due to the limitations in soft body dynamics, tactile imprints generated by the simulator are reasonably similar to that of the real-world. Figure \ref{fig:real_vs_sim} shows a comparison of the simulated and real tactile images from the same objects.

The simulator is made publicly available at: \newline 
{\scriptsize \texttt{https://github.com/SAIC-MONTREAL/multimodal-dynamics}}

\section{Data Collection Details}
Supplementary video for simulation and real-world data collection is available at: \newline
{\scriptsize \texttt{https://sites.google.com/view/multimodal-dynamics}}

\section{Supplementary Prediction Results}
In this section we show more prediction results from the three simulated scenario in Fig. \ref{fig:sim_res_falling_large}, \ref{fig:sim_res_sloped_large}, and \ref{fig:sim_res_shock_large} to supplement those in the paper.

{\onecolumn
\begin{sidewaysfigure}[b!]
    \centering
    \includegraphics[width=1.05\textwidth]{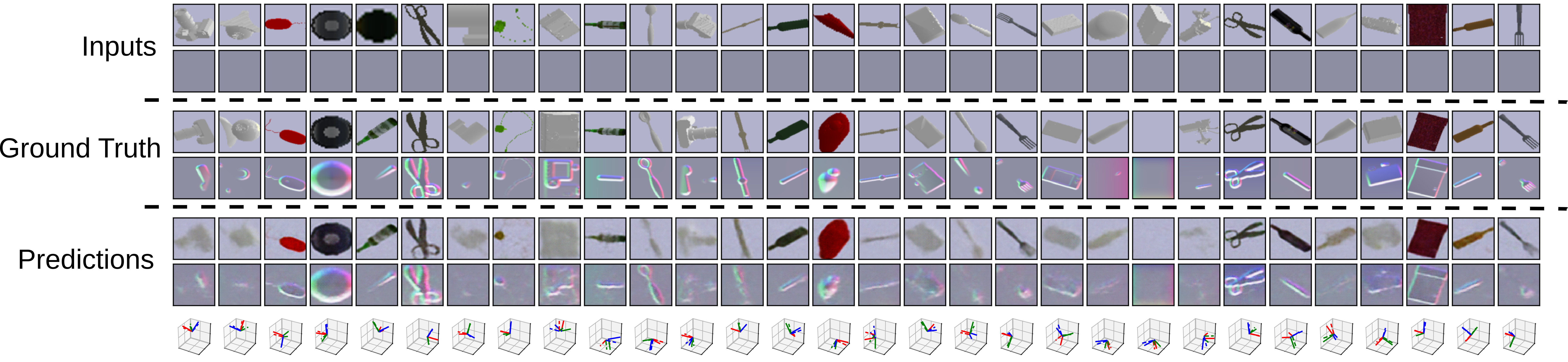}
    \caption{Multimodal predictions for the simulated freefalling scenario evaluated on the validation set. The motion starts from a non-contacting initial position, accounting for the initial unavailability of  tactile measurements. The model predicts the final resting pose in addition to the visual and tactile measurements of the STS sensor. The bottom row compares the predicted pose (solid coordinates) with the ground truth (dashed coordinates).}
    \label{fig:sim_res_falling_large}
\end{sidewaysfigure}

\begin{sidewaysfigure}[b!]
    \centering
    \includegraphics[width=1.05\textwidth]{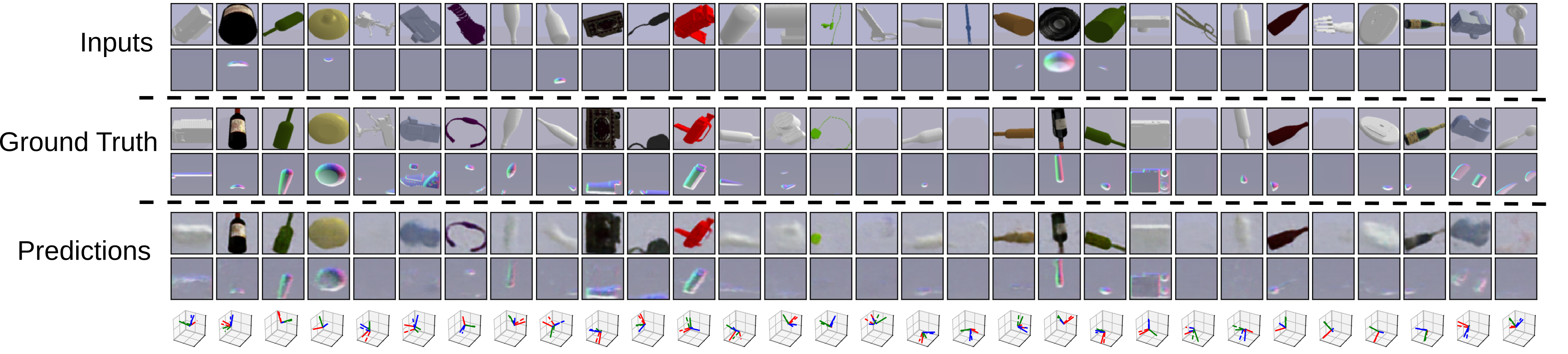}
    \caption{Multimodal predictions for the simulated inclined surface scenario evaluated on the validation set. The motion starts in proximity to the surface, accounting for the the initially availability of tactile measurements that may be either active or inactive. The model predicts the final resting pose in addition to the visual and tactile measurements of the STS sensor. The bottom row compares the predicted pose (solid coordinates) with the ground truth (dashed coordinates).}
    \label{fig:sim_res_sloped_large}
\end{sidewaysfigure}

\begin{sidewaysfigure}[b!]
    \centering
    \includegraphics[width=1.05\textwidth]{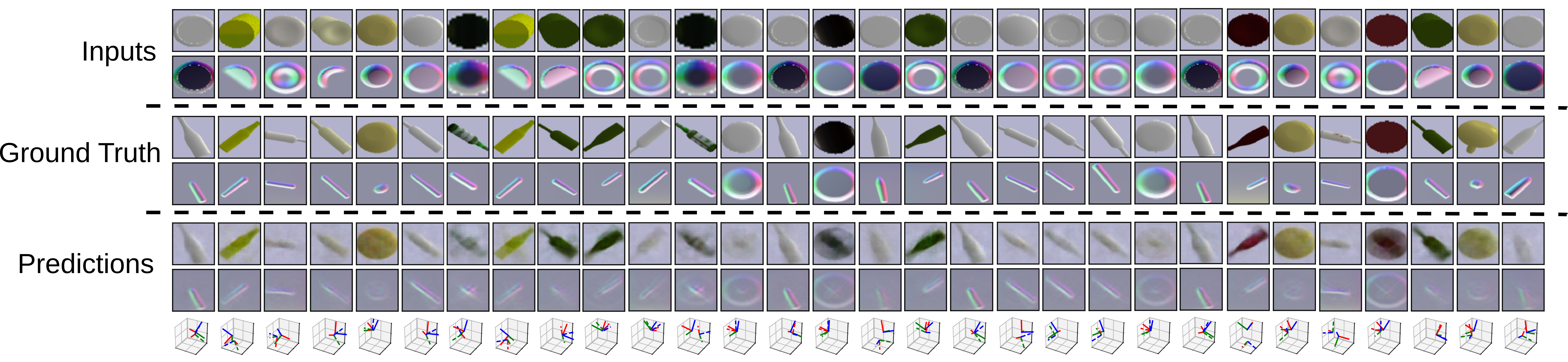}
    \caption{Multimodal predictions for the simulated force perturbation scenario evaluated on the validation set.  The bottle starts in an initial stable resting state, accounting for the initial availability of tactile measurements. The model predicts the final resting pose in addition to the visual and tactile measurements of the STS sensor. The bottom row compares the predicted pose (solid coordinates) with the ground truth (dashed coordinates).}
    \label{fig:sim_res_shock_large}
\end{sidewaysfigure}
}

\twocolumn

\end{document}